\documentclass[lettersize,journal]{IEEEtran}
\usepackage{amsmath,amsfonts}
\usepackage{array}
\usepackage[numbers,sort&compress]{natbib}
\usepackage[caption=false,font=normalsize,labelfont=sf,textfont=sf]{subfig}
\usepackage{textcomp}
\usepackage{stfloats}
\usepackage{url}
\usepackage{verbatim}
\usepackage{graphicx}
\usepackage[utf8]{inputenc} 
\usepackage[T1]{fontenc}    
\usepackage{hyperref}       
\hypersetup{
    colorlinks=true,       
    linkcolor=black,       
    citecolor=black,       
    urlcolor=black         
}
\usepackage{url}            
\usepackage{booktabs}       
\usepackage{nicefrac}       
\usepackage{microtype}      
\usepackage{xcolor}         
\usepackage{natbib}
\setcitestyle{numbers,square}
\usepackage{framed,multirow}
\usepackage{latexsym}
\usepackage{amsmath,amssymb,amsfonts}
\usepackage{algorithmic}
\usepackage{algorithm}
\usepackage{multirow}
\usepackage{longtable}
\usepackage{ragged2e}
\usepackage{pifont}
\usepackage{textcomp}
\usepackage{makecell}
\usepackage[figuresright]{rotating}
\usepackage{colortbl}
\definecolor{mygray}{gray}{.9}

\def\BibTeX{{\rm B\kern-.05em{\sc i\kern-.025em b}\kern-.08em
    T\kern-.1667em\lower.7ex\hbox{E}\kern-.125emX}}
\usepackage{balance}

\begin{document}
\title{Uncertainty-aware Medical Diagnostic Phrase Identification and Grounding}
\author{Ke Zou, Yang Bai, Bo Liu, Yidi Chen, Zhihao Chen, Yang Zhou, Xuedong Yuan, Meng Wang, Xiaojing Shen,\\ Xiaochun Cao~\IEEEmembership{Senior Member, IEEE}, Yih Chung Tham, Huazhu Fu~\IEEEmembership{Senior Member, IEEE}
\thanks{Manuscript received December, 2024; revised June, 2025. This work was supported by the National Key Researchand Development Program of China under Grant 2020YFA0714003, the National Natural Science Foundation of China (No.62025604, 62411540034), the H. Fu’s Agency for Science, Technology and Research (A*STAR) Central Research Fund (“Robust and Trustworthy AI system for Multi-modality Healthcare”), the Science and Technology Department of Sichuan Province (Grant No. 2022YFS0071), the Science and Technology Department of Guangxi Zhuang Autonomous Region (Grant No. 2025GXNSFAA069531), Science and Technology Department of Hainan Province (Grant number ZDYF2024SHFZ052) and the China Scholarship Council (No. 202206240082).}
\thanks{K. Zou and X. Yuan are with the College of Computer Science, Sichuan University, Chengdu, China. }
\thanks{Y. Bai, Y. Zhou, M. Wang, and H. Fu are with the Institute of High Performance Computing (IHPC), Agency for Science, Technology and Research (A*STAR), Singapore. }
\thanks{Bo Liu is with the Department of Computing, The Hong Kong Polytechnic University, Hong Kong, China.}
\thanks{Y. Chen is with the Department of Radiology, West China Hospital, Sichuan University, Chengdu, China. }
\thanks{Z. Chen is with the College of Intelligence and Computing, Tianjin University, Tianjin, China }
\thanks{X. Shen is with the Department of Mathematics, Sichuan University, Chengdu, China. }
\thanks{X. Cao is with the School of Cyber Science and Technology, Shenzhen Campus of Sun Yat-Sen University, Shenzhen, China. }
\thanks{K. Zou and Y. Tham are with the Department of Ophthalmology, Yong Loo Lin School of Medicine, National University of Singapore and the Singapore Eye Research Institute, Singapore National Eye Centre, Singapore, Singapore.}
\thanks{Ke Zou and Yang Bai are equally contributed. Xuedong Yuan and Huazhu Fu are the co-corresponding authors (e-mail: yxdongdong@163.com, hzfu@ieee.org).}}

\markboth{Journal of \LaTeX\ Class Files,~Vol.~18, No.~9, September~2020}%
{How to Use the IEEEtran \LaTeX \ Templates}

\maketitle

\begin{abstract}
Medical phrase grounding is crucial for identifying relevant regions in medical images based on phrase queries, facilitating accurate image analysis and diagnosis. However, current methods rely on manual extraction of key phrases from medical reports, reducing efficiency and increasing the workload for clinicians.  Additionally, the lack of model confidence estimation limits clinical trust and usability. In this paper, we introduce a novel task called Medical Report Grounding (MRG), which aims to directly identify diagnostic phrases and their corresponding grounding boxes from medical reports in an end-to-end manner. To address this challenge, we propose uMedGround, a robust and reliable framework that leverages a multimodal large language model to predict diagnostic phrases by embedding a unique token, $<$$\mathtt{BOX}$$>$, into the vocabulary to enhance detection capabilities. A vision encoder-decoder processes the embedded token and input image to generate grounding boxes. Critically, uMedGround incorporates an uncertainty-aware prediction model, significantly improving the robustness and reliability of grounding predictions. Experimental results demonstrate that uMedGround outperforms state-of-the-art medical phrase grounding methods and fine-tuned large visual-language models, validating its effectiveness and reliability. This study represents a pioneering exploration of the MRG task, marking the first-ever endeavor in this domain. Additionally, we demonstrate the applicability of uMedGround in medical visual question answering and class-based localization tasks, where it highlights visual evidence aligned with key diagnostic phrases, supporting clinicians in interpreting various types of textual inputs, including free-text reports, visual question answering queries, and class labels.
\end{abstract}

\begin{IEEEkeywords}
Medical report grounding, vision-language model, uncertainty estimation.
\end{IEEEkeywords}

\begin{figure}[!t]
\centering
\includegraphics[width=1\linewidth]{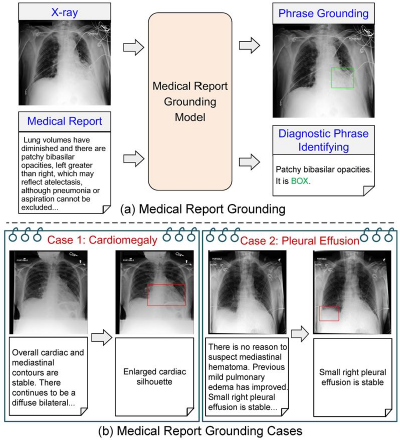}
\caption{ Medical report grounding. (a) Illustration of our medical report grounding with large-language model for radiological diagnosis. (b) Different paired cases (X-ray image, medical report, grounding box and key phrase): Cardiomegaly and Pleural Effusion.}
\label{F_1}
\end{figure}

\section{Introduction}
Medical image grounding involves detecting bounding boxes of corresponding categories from medical images, assisting clinicians in identifying abnormalities. To facilitate a more direct approach for localizing areas that clinicians need to sketch based on diagnostic phrases, Chen et al.~\cite{chen2023} introduced the medical phrase grounding. However, this task requires first extracting the relevant phrase from a medical report and then detecting the corresponding region in the image. In this paper, we propose a novel task, \textbf{M}edical \textbf{R}eport \textbf{G}rounding (\textbf{MRG}), which directly identifies diagnostic phrase and their corresponding grounding box from the medical report. As illustrated in Fig.~\ref{F_1} (a), this task presents two key challenges: identifying diagnostic phrase within the report and grounding bounding box associated with it. Fig.~\ref{F_1} (b) provides additional examples using X-ray imaging, showcasing cases such as cardiac hypertrophy and pleural effusion diagnoses. This process plays a crucial role in enhancing the comprehension of semantic information within medical reports, enabling the identification of grounding box associated with phrase in medical image.

In computer vision, tasks similar to MRG include visual grounding, which has been extensively explored in the context of natural images~\cite{deng2021transvg,vgtr2022,zhu2022seqtr,zhou2023joint,he2024improved}. In contrast, its application to medical imaging remains limited, primarily due to challenges in labeling and acquiring medical data. Chen et al.~\cite{chen2023} and Ichinose et al~\cite{Ichinose2023} first introduced methods for medical phrase X-ray grounding and medical report CT grounding. However, their approaches rely heavily on pre-processing steps to extract diagnostic phrases from medical reports, which limits system efficiency and scalability. Recent advances in large vision-language models (VLMs), such as GPT-4V~\cite{achiam2023gpt}, LLaVA-13B~\cite{liu2024llava}, and InternVL~\cite{chen2024internvl,chen2024far}, have shown strong capabilities in multi-modal understanding and reasoning. However, the precision and reliability of these models in identifying diagnostic phrase and grounding corresponding bounding box remain to be fully optimized. These models combine large language models (LLMs) with vision encoders to directly generate textual responses from input images. While promising, their ability to precisely identify diagnostic phrases and localize corresponding regions with bounding boxes remains suboptimal, especially in clinical deployment scenarios where interpretability and reliability are paramount.

To this end, an important but often overlooked aspect is the uncertainty associated with model predictions, especially bounding box predictions in medical contexts. Unlike natural image tasks, vague or overly confident localization in clinical settings could lead to critical misinterpretations. Uncertainty estimation offers a potential solution by transforming single-point predictions into probabilistic or multi-hypothesis representations. Common uncertainty estimation methods in medical domain include Monte-Carlo Dropout (MCDO)~\cite{MIA2020exploringDropSeg}, Ensemble~\cite{UAMT2019ensembleSeg}, Test-time data augmentation (TTA)~\cite{wang2018TTA}, evidential-based~\cite{huang2022lymphoma,zou2022tbrats,wang2023uncertainty} and deterministic-based methods~\cite{evidential18,2021DetermSeg,ICMLdeterministic20}. However, such techniques have rarely been adapted to phrase grounding, and direct integration into large VLMs may incur prohibitive computational costs. To address these limitations, we propose a lightweight and scalable multi-hypothesis prediction framework, which serves as a plug-and-play module within vision-language models. This design improves grounding robustness without significantly increasing inference time or model complexity. Importantly, it provides variance-aware outputs, offering interpretable visual cues that can enhance trust and adoption in clinical workflows.

In this paper, we introduce a new task, medical report grounding, which addresses two key challenges: identifying diagnostic phrase and grounding their corresponding box. To tackle these challenges, we propose an innovative framework \textbf{u}ncertainty-aware \textbf{Me}dical \textbf{d}iagnostic phrase identification and \textbf{ground}ing, named \textbf{uMedGround}. It begins by leveraging a multimodal LLM to enhance the understanding and interpretation of medical report, enabling accurate identification of diagnostic phrase. Once the multimodal LLM identifies the diagnostic phrase, a novel token, $<\mathtt{BOX}>$ is embedded into the vocabulary to represent the phrase prediction. Then, the implicit embedding of $<\mathtt{BOX}>$, together with the input medical image, is then decoded using a SAM-based encoder-decoder to generate the corresponding grounding box. Most critically, we introduce an uncertainty-aware prediction model to quantify confidence in the grounding box, enhancing both robustness and reliability. This is achieved by integrating a multi-hypothesis prediction head into the fine-tuning process of the box decoder, enabling the simultaneous output of the box and its associated uncertainty. Our contribution can be summarized as follows\\
$\bullet$ We propose a novel task, medical report grounding, which involves diagnostic phrase identification and grounding box prediction. This task assists doctors in identifying key diagnostic information and its visual cues within medical reports and images, facilitating improved diagnostic accuracy and supporting more informed clinical decision-making.\\
$\bullet$ An end-to-end framework is developed in this study, uMedGround, which incorporates a reliable multimodal LLM to enhance the understanding and interpretation required for medical report grounding.\\
$\bullet$ A new $<$$\mathtt{BOX}$$>$ token is introduced as an embedding input to the box decoder, significantly improving the precision of critical finding localization within medical images.\\
$\bullet$ We pioneer the integration of uncertainty estimation into medical report grounding, enhancing both the accuracy and reliability of grounding predictions.\\
$\bullet$ Three publicly available datasets are preprocessed and curated to generate four paired subdatasets: two designed for experimental validation and two intended for clinical application testing. Comprehensive experiments and applications on these datasets demonstrate the accuracy, reliability, and applicability of the proposed framework. This study represents a pioneering exploration of the MRG task, marking the first-ever endeavor in this domain.\footnote{Our codes and curated datasets will be released in https://github.com/Cocofeat/uMedGround.}

\section{Related works}
\label{S_2}
\subsection{Visual-language model for grounding}
Visual grounding had been extensively studied in the context of natural images, evolving from early two-stage models~\cite{plummer2018conditional,wang2019phrase} to multi-modal transformer-based approaches~\cite{vgtr2022, deng2021transvg, zhu2022seqtr,zhou2023joint,he2024improved}. However, these methods faced challenges when applied to radiological images. MRG demanded specialized visual-textual feature learning to enable the model to discern medical findings that exhibited subtle textural and morphological variations and to interpret the spatial relations described in medical reports. The transition from general visual grounding to the more specialized domain of MRG underscored the need for tailored approaches.

This need was reflected in several pivotal studies. Moon et al.~\cite{moon2022multi} leveraged large-scale medical image-text data for vision-language pretraining, highlighting the potential of AI in understanding complex multimodal medical narratives. Wang et al.~\cite{wang2022multi} proposed a cross-modal contrastive attention model that bridged the gap between radiology images and reports through aligned multi-modal representations. Despite these advancements in knowledge representation, a performance gap remained in translating this understanding into accurate and clinically valid grounding predictions. More recently, large vision-language models (VLMs) such as GPT-4V~\cite{achiam2023gpt}, LLaVA-13B~\cite{liu2024llava}, and InternVL~\cite{chen2024internvl,chen2024far} demonstrated significant potential for multimodal understanding and reasoning. In the medical domain, visual-language pretraining frameworks like MedKLIP~\cite{wu2023medklip} and BioViL~\cite{BioViL2022} incorporated clinical knowledge to improve grounding performance, enabling fine-grained contextual understanding critical for downstream tasks. Among early MRG efforts, MedRPG~\cite{chen2023} adopted a transformer-based model to predict bounding boxes given pre-extracted medical phrases, while Ichinose et al.~\cite{Ichinose2023} introduced a framework for grounding in 3D CT-report pairs by structuring reports to guide phrase localization. Despite their effectiveness, these methods required explicit report preprocessing to isolate key phrases, which limited their scalability and adaptability.

To overcome these limitations, we proposed a method that directly extracted key phrases from free-text reports using LLMs, allowing for intuitive understanding and seamless end-to-end grounding. In a similar direction, Vilouras et al.~\cite{vilouras2024zero} employed a generative foundation model for zero-shot medical phrase grounding and achieved competitive results. While large-scale VLMs for medicine remained underexplored, existing work primarily targeted tasks such as disease classification, visual question answering (VQA), and radiology report generation~\cite{wang2023chatcad, moor2023med, wu2023towards}. To address the long-tail distribution of clinical knowledge and enhance diagnostic reasoning, recent studies integrated heterogeneous external sources, including multi-source retrieval~\cite{zhao2024heterogeneous} and medical knowledge graphs~\cite{gao2023large}, into LLM-based QA systems. MARIA-2~\cite{bannur2024maira} advanced radiology report generation by evaluating accuracy and completeness, although it remained task-specific. In contrast, our model aimed to deliver a unified and generalizable solution for grounding that could also be extended to classification-based localization and VQA tasks.

\subsection{Uncertainty estimation in medical image analysis}
Uncertainty estimation had become an increasingly important topic in medical image analysis, serving as a means for improving reliability and interpretability. According to Huang et al.~\cite{huang2024review}, uncertainty estimation had been explored through both probabilistic~\cite{2018probabilisticU,MIA2020exploringDropSeg,TMI20ensembleSeg,wang2018TTA,luo2024measurement} and non-probabilistic~\cite{zou2022tbrats,wang2023uncertainty,huang2025deep} approaches across various medical imaging tasks.

For probabilistic methods, Kohl et al.~\cite{2018probabilisticU} introduced a probabilistic U-Net for segmenting ambiguous medical images, though its computational cost was prohibitively high. Later, Nair et al.~\cite{MIA2020exploringDropSeg} investigated MC dropout and a range of uncertainty metrics in deep learning for multiple sclerosis lesion segmentation. Ensemble learning also emerged as a viable approach; Mehrtash et al.~\cite{TMI20ensembleSeg} used ensembles to improve confidence calibration and predictive reliability in deep segmentation networks.

On the non-probabilistic side, Wang et al.~\cite{wang2018TTA} proposed a brain tumor segmentation method that utilized test-time augmentation to estimate uncertainty and boost robustness. However, these techniques typically required multiple inferences per image and thus introduced additional computational burdens. To address this limitation, Zou et al.~\cite{zou2022tbrats} and Wang et al.~\cite{wang2023uncertainty} applied evidential deep learning to brain tumor and retinal disease segmentation, offering single-pass uncertainty estimation that improved both robustness and generalizability. In summary, although uncertainty estimation methods have seen significant application in medical classification and segmentation, their integration into medical grounding tasks is still limited. Furthermore, directly incorporating these methods into medical grounding tasks may significantly increase model parameters and inference time, particularly for VLMs.

\section{Methodology \label{S_4}}
\subsection{Problem Definition\label{S_3}}
Medical report grounding aims to extract crucial diagnostic information, resembling the expertise of an X-ray specialist, from an original medical report ${{\bf{x}}_t}$ and an X-ray image ${{\bf{x}}_i}$. Specifically, the objective is to identify the medical phrase ${{\bf{y}}_p}$ and its corresponding detection grounding box ${{\bf{y}}_b}$ within the X-ray image. This process can be summarized as follows:
\begin{equation}
\label{E_1}
\left( {{{\bf{y}}_p},{{\bf{y}}_b}} \right) = \mathcal{F}\left( {{{\bf{x}}_i},{{\bf{x}}_t}} \right),
\end{equation}
$b = (x, y, w, h)$, $(x, y)$ denote the center coordinates of the bounding box in the image, while $(w, h)$ correspond to its height and width, respectively. Furthermore, $ \mathcal{F}\left(  \cdot  \right)$ represents the mapping function that relates the input variables to the output. While Chen et al.~\cite{chen2023} and Ichinose et al~\cite{Ichinose2023} proposed medical report grounding, their approaches necessitate manual preprocessing of the phrases. Leveraging the reasoning capabilities of large VLMs, our method automatically infers phrases from original medical reports and conducts corresponding grounding box. This advancement significantly enhances the efficiency of medical personnel.

Next, we first provide an overview of our architecture for medical phrase prediction and grounding box detection. We then review the multi-hypothesis prediction model and build upon it to establish an uncertainty-aware prediction model for grounding boxes. Furthermore, we present more details of the uncertainty-aware box decoder. Finally, we introduce the overall training loss function.

\subsection{The Architecture of Medical Report Grounding \label{4_1}}
\textbf{Medical phrase identification from medical report.} Our primary objective is to generate medical phrases as input for predicting grounding boxes. However, being restricted solely to medical phrases may lead to unnecessarily complicating the framework as in~\cite{chen2023}. The currently booming large-scale visual language models can predict text by taking images and text as output. Remarkably, building upon the insights from~\cite{lai2023lisa}, we leverage a large visual language model to generate phrases and utilize their embeddings as inputs for predicting grounding boxes, thereby enhancing the elegance of the framework. Consequently, given an X-ray image ${{\bf{x}}_i}$ and a medical report ${{\bf{x}}_t}$, we first obtain the objective function as follows:
\begin{equation}
\label{E_2}
\left\{ {\begin{array}{*{20}{c}}
{{{\bf{e}}} = {\mathcal F_{vlm}}\left( {{{\bf{x}}_i},{{\bf{x}}_t}} \right)}\\
{{{\bf{y}}_p} = {f_\theta }({\bf{e}})}
\end{array}} \right..
\end{equation}
Here, \( \mathcal{F}_{\text{vlm}} \) denotes the large VLM equipped with LoRA~\citep{hu2022lora} for parameter-efficient fine-tuning. LoRA adapters are inserted into the attention modules of the VLM backbone, specifically targeting the query and value projection layers with a rank \( r = 8 \) and a scaling factor \( \alpha = 16 \), while the original model weights are frozen and only the LoRA parameters are updated during training. The function \( {f_\theta }( \cdot )\) represents the MLP projection layer, which predicts the phrase \( \mathbf{y}_p \) based on the phrase prediction embedding \( \mathbf{e} \). It is noteworthy that in Eq.~\ref{E_2}, we employ the box embedding ${{\bf{e}}_{box}}$ in the ${\bf{e}}$ as a detection paradigm, integrating detection capabilities into the VLM. To facilitate this, we expand the vocabulary of the original language model by a new token, namely $<$$\mathtt{BOX}$$>$, which serves as a marker for requesting detection box outputs. Next, we utilize the $<$$\mathtt{BOX}$$>$ token corresponding to the final-layer embedding of the VLM to predict the grounding box.

\textbf{Phrase Grounding.} First, we can directly use LLaVA equipped with LoRA to predict the initial grounding box \( {\bf{y}}_b^0 \), that is:
\begin{equation}
\label{E_3_pre}
{\bf{y}}_b^0{\rm{ = }}{f_\Phi }\left( {\bf{e}}_{box} \right),
\end{equation}
This indicates that the $<$$\mathtt{BOX}$$>$ token is directly used as input to an MLP layer ${f_\Phi }\left( \cdot \right)$ for grounding box prediction. However, this grounding box may not effectively leverage the information from the box embedding ${{\bf{e}}_{box}}$.
Therefore, to enhance the accuracy of detection grounding box, we adopt the widely-used SAM framework~\cite{SAM}, utilizing the $<$$\mathtt{BOX}$$>$ embedding as a prompt for precise prediction. Specifically, we employ the visual backbone ${\mathcal F_{enc}}$ to extract visual features ${{\bf{z}}_{enc}}$. The box embedding ${\bf{e_{box}}}$ is then combined with the visual features and fed into the decoder ${\mathcal F_{dec}}$. Notably, ${\bf{e_{box}}}$ serves as a contextual cue. Compared with SAM framework~\cite{SAM}, we enhance our model by modifying the final layer to box decoder, thereby obtaining grounding box embedding ${{\bf{z}}_{dec}}$. The process can be described as follows:
\begin{equation}
\label{E_3}
\left\{ {\begin{array}{*{20}{c}}
{{{\bf{z}}_{enc}} = {\mathcal{F}_{enc}}\left( {{{\bf{x}}_i}} \right)}\\
{{{\bf{z}}_{dec}} = {\mathcal{F}_{dec}}\left( {{{\bf{z}}_{enc}},{{\bf{e}}_{box}}} \right)}
\end{array}} \right..
\end{equation}

\begin{figure*}[!t]
\centering
\includegraphics[width=1\linewidth]{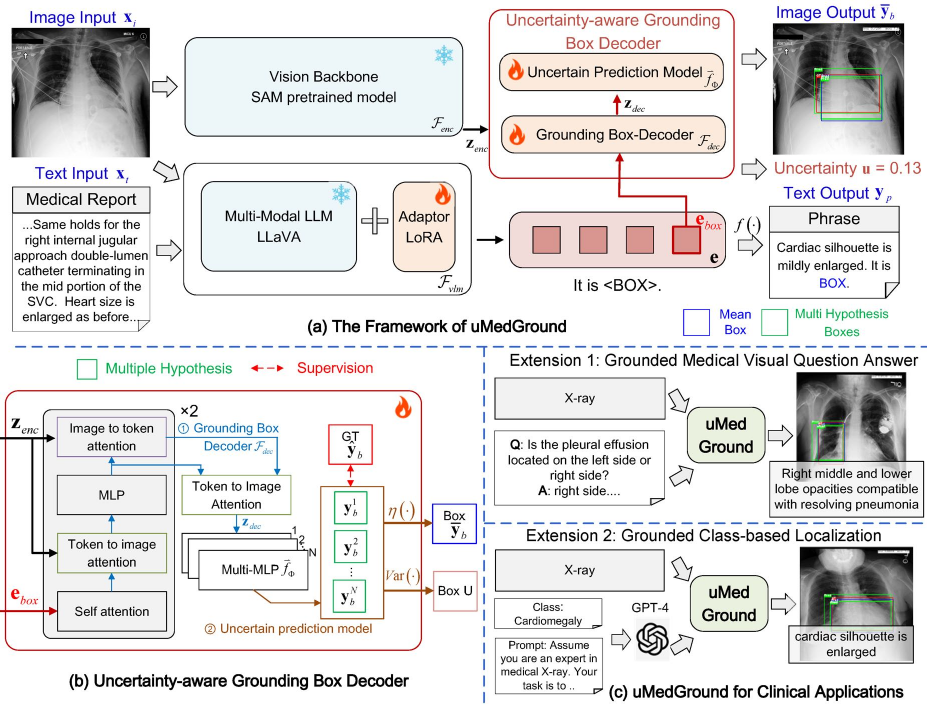}
\caption{Overview of \textbf{uMedGround}, an uncertainty-aware vision-language model designed for identifying and grounding medical diagnostic phrase. (a) The overall framework of uMedGround, which aligns medical reports with uncertainty-aware visual evidence at the phrase level. (b) The uncertainty-aware box decoder, which refines spatial grounding predictions based on multiple hypothesis predictions. (c) A potential clinical extension demonstrating how uMedGround can assist in explainable decision-making by linking grounded phrases with corresponding image regions.}
\label{F_2}
\end{figure*}

\subsection{Multiple Hypothesis Prediction Model\label{4_b2}}
To enhance the reliability of ground box predictions, we transition the prediction model from a single-point estimation framework to a multi-hypothesis estimation paradigm. In this context, we provide a brief review of multi-hypothesis prediction models. We assume the development of a prediction function capable of generating $N$ predictions for a given input $x$:
\begin{equation}
\label{E_b4}
{{\mathord{\buildrel{\lower3pt\hbox{$\scriptscriptstyle\rightharpoonup$}} 
\over f} }_\Phi }\left( x \right) = \left( {f_\Phi ^1\left( x \right), \ldots ,f_\Phi ^N\left( x \right)} \right),
\end{equation}
Where ${{\mathord{\buildrel{\lower3pt\hbox{$\scriptscriptstyle\rightharpoonup$}} 
\over f} }_\Phi }:{\cal X} \to {\cal Y}$ is a predictor, $x \in \mathcal{X}$ and $y \in \mathcal{Y}$ as the input and output variables. Then, calculating the loss $\mathcal{L}$ as always being the closest of the $N$ predictions will minimize the following objective:
\begin{equation}
\label{E_b5}
\int_{\mathcal{X}} {\sum\limits_{i = 1}^N {\int_{{\mathcal{Y}_i}\left( x \right)} {\rm{ }} } } (f_\Phi ^i(x),y)p(x,y)\;dy\;dx,
\end{equation}
$p(x,y) = p(y|x)p(x)$ denotes the joint probability density of the input variable $x$ and the label $y$. We consider the Voronoi tessellation of the label space \( \mathcal{Y} = \bigcup_{i=1}^N \mathcal{Y}_i \), which is induced by \( M \) generators \( g_(x) \), and the loss function \( \mathcal{L} \):
\begin{equation}
\label{E_b6}
{{\mathcal{Y}}_i}(x) = \left\{ {y \in {\mathcal{Y}}:{\rm{ }}({g^i}(x),y) < ({g^j}(x),y)\;\forall j \ne i} \right\}.
\end{equation}
The Voronoi tessellation can be intuitively understood as partitioning the space such that each cell contains all points that are closest to a specific generator. In this case, the notion of closeness is determined by the loss function \( \mathcal{L} \). Therefore, Eq.~\ref{E_b5} partitions the space into \( N \) Voronoi cells, each corresponding to one of the predicted hypotheses \( f_i(x) \), and aggregates the loss associated with each cell.

\textbf{Theorem 1} For Eq.~\ref{E_b5} to attain its minimum, a necessary condition is that the generators \( g^j(x) \) coincide with the predictors \( f^j_\theta(x) \), and both must correspond to a centroidal Voronoi tessellation:
\begin{equation}
\label{E_b6}
{g^i}(x) = f_\Phi ^i(x) = \frac{{\int_{{\mathcal{Y}_i}} \mathcal{L} (f_\Phi ^i(x),y)p(y|x)\;dy}}{{\int_{{\mathcal{Y}_i}} p (y|x)\;dy}}.
\end{equation}
For a more detailed proof and derivation of this theorem can be found in~\cite{du1999centroidal}. Intuitively, minimizing Eq.~\ref{E_b5} aims to find the best piecewise constant approximation to the conditional distribution of labels in the output space. The hypotheses will divide the space into cells in such a way that the expected loss relative to their conditional means is minimized. Next, we will further explain the proposed uncertain prediction model by incorporating the multi-hypothesis prediction model.

\subsection{Uncertain Prediction Model for Grounding Box  \label{4_2}}
Directly predicting the grounding box using \( \mathbf{z}_{\text{dec}} \) from Eq.~\ref{E_3} often results in a single deterministic output, which can suffer from overconfidence or insufficient reliability, making it difficult to gain acceptance from clinical experts in medical diagnostics. To address this limitation, we propose the  \textbf{U}ncertain \textbf{P}rediction \textbf{M}odel (\textbf{UPM}). UPM incorporates the multi-hypothesis model described earlier, utilizing multiple hypothesis predictions to maintain output diversity, rather than relying solely on a single deterministic prediction model. More importantly, our model reveals the uncertainty (i.e., variance) of multiple hypothesis predictions, thereby overcoming unrealistic or blurred predictions in regression grounding box prediction. By incorporating this model into the vision output of LLMs, we provide a more reliable and robust grounding box prediction, which is crucial for gaining the trust and confidence of clinical experts in the field of medical diagnosis.

Specifically, instead of the expected value distribution for a single prediction ${{\bf{y}}_b}{\rm{ = }}{f_\Phi }\left( {{{\bf{z}}_{dec}}} \right)$, we aim to obtain multiple predictions to potentially achieve better approximations. To this end, we assume that the grounding box prediction is provided by $N$ predictions:
\begin{equation}
\label{E_4}
{{\mathord{\buildrel{\lower3pt\hbox{$\scriptscriptstyle\rightharpoonup$}}
\over f} }_\Phi }\left( {{{\bf{z}}_{dec}}} \right) = \left( {f_{_\Phi }^1\left( {{{\bf{z}}_{dec}}} \right), \ldots ,f_{_\Phi }^N\left( {{{\bf{z}}_{dec}}} \right)} \right),
\end{equation}
where ${f_{_\Phi }^N\left( {{{\bf{z}}_{dec}}} \right)}$ represents the $N$-th  prediction hypothesis derived from the embedding feature ${\bf{z}}_{dec}$. Next, similar to Eq.~\ref{E_b5}, we calculate that the uncertain-aware grounding box prediction loss ${{\cal L}_u}$ that is always the closest one out of $N$ predictions, thus recommending minimization:
\begin{equation}
\label{E_41}
\begin{array}{l}
\sum\limits_{i = 1}^N {{\mathcal{L}_u}\left( {{{{\bf{\hat y}}}_b},f_{_\Phi }^i\left( {{{\bf{z}}_{dec}}} \right)} \right)}, 
\end{array}
\end{equation}
Then, we consider the Voronoi tessellation of the label space ${{\bf{y}}_b} =  \cup _{i = 1}^N{\bf{y}}_b^i$, which consists of multiple hypothesis predictions. Subsequently, we use the mean among the multiple hypothesis predictions as the final prediction and the variance as the uncertainty, which can be expressed as follows: 
\begin{equation}
\label{E_5}
{{{\bf{\bar y}}}_b} = \eta \left( {\sum\limits_{i = 1}^N {{\bf{y}}_b^i} } \right),{\rm{  }}{\bf{u}} = Var\left( {\sum\limits_{i = 1}^N {{\bf{y}}_b^i} } \right),
\end{equation}
where $\eta \left(  \cdot  \right)$ and $Var \left(  \cdot  \right)$ denote the mean and variance of their predictions. 

\subsection{Uncertainty-aware Grounding Box Decoder}
To obtain uncertainty estimation for grounding box prediction, we introduce the uncertainty-aware grounding box decoder, detailed in Fig.~\ref{F_2} (b). Comprising two essential components, namely the grounding box decoder ${\cal F}_{dec}$ as described in Sec.~\ref{4_1} and the UPM $\mathop \sum \limits_{i = 1}^N f_\Phi ^i$ detailed in Sec.~\ref{4_2}, this decoder primarily decodes the input image embedding and box embedding, producing a multi-modal feature embedding ${\bf{z}}_{dec}$. Additionally, UPM is employed for forecasting supervision and uncertainty estimation of joint embedding. Specifically, we employ multiple MLP layers $\mathord{\buildrel{\lower3pt\hbox{$\scriptscriptstyle\rightharpoonup$}} 
\over f} \left(  \cdot  \right) = {\left( {{f_1}\left(  \cdot  \right),{f_2}\left(  \cdot  \right), \ldots ,{f_N}\left(  \cdot  \right)} \right)}$, to obtain multiple hypothesis predictions ${\sum\limits_{i = 1}^N {{\bf{y}}_b^i} }$ of the grounding boxes. In the end, the parameters of the box distribution are employed in Eq.~\ref{E_5} to compute the predictions for both the grounding box and uncertainty.
Therefore, the final expression of Eq.~\ref{E_1} can be represented as:
\begin{equation}
\label{E_7}
\left( {{{\bf{y}}_p},{{{\bf{\bar y}}}_b},\bf{u}} \right) = {\cal F}\left( {{{\bf{x}}_t},{{\bf{x}}_i}} \right).
\end{equation}


\subsection{Training to Form Medical Report Grounding}
The proposed model is designed to be trained end-to-end and consists of three key components: phrase generation, detection box prediction, and uncertainty estimation.\\
\textbf{Phrase identification.} To enhance the ability of the visual language model to effectively summarize medical phrases in medical reports, we employ an auto-regressive cross-entropy loss. This loss function enables the model to learn the correlation between the predicted phrase and the actual phrase. Concretely, during the training process, given a real-valued phrase ${{{\bf{\hat y}}}_p}$, the formulation of the training process is as follows:
\begin{equation}
\label{E_8}
{\mathcal {L}_p}{\rm{ = }}{\mathcal {L}_{CE}}\left( {{{{\bf{\hat y}}}_p},{{\bf{y}}_p}} \right).
\end{equation}
\textbf{Coarse phrase grounding.} Furthermore, we supervise the initial prediction of the grounding box in Eq.~\ref{E_2} using two loss functions: smooth L1 loss~\cite{fastrcnn2015} and generalized intersection over union (GIoU) loss~\cite{giou2019}. These loss functions serve as guidance for improved box prediction. Specifically, during the optimization process, given a real box ${{{\bf{\hat y}}}_b}$, the formulation can be expressed as follows:
\begin{equation}
\label{E_9}
{{\cal L}_b}{\rm{ = }}{{\cal L}_{{l_1}}}\left( {{{\bf{\hat y}}}_b},{{{\bf{y}}_b^0}} \right) + {{\cal L}_{giou}}\left( {{{{\bf{\hat y}}}_b},{{{\bf{y}}}_b^0}} \right).
\end{equation}
\textbf{Uncertainty-aware phrase grounding.} As indicated in Eq.~\ref{E_41}, we utilize the uncertainty-aware prediction model to perform the final grounding box prediction. This optimization process combines the aforementioned L1 loss and GIoU loss, leading to the following equation:
\begin{equation}
\label{E_10}
{{\cal L}_u} = \sum\limits_{i = 1}^N {{{\cal L}_{{l_1}}}\left( {{{{\bf{\hat y}}}_b},{w_i}{\bf{y}}_b^i} \right)}  + {{\cal L}_{giou}}\left( {{{{\bf{\hat y}}}_b},{w_i}{\bf{y}}_b^i} \right).
\end{equation}
Unlike Eq.~\ref{E_41}, we introduce a weight \( w_i \) to relax the hard minimization constraint in it. This idea stems from the fact highlighted in~\cite{rupprecht2017Mheads}: the predictor ${f_\Phi }\left(  \cdot  \right)$ might be initialized far from the target label ${\bf{\hat y}}_b$, such that all \( y \) fall into a single Voronoi cell. The weights are assigned following these principles:
\begin{equation}
\label{E_10}
w(\alpha) =
\begin{cases} 
1 - \lambda, & \text{if } \alpha \text{ is true}, \\
\frac{\lambda}{N - 1}, & \text{otherwise}.
\end{cases}
\end{equation}
The label \( y_b \) is assigned to the nearest hypothesis with a weight of \( 1 - \lambda \), while the remaining hypotheses share a weight of \( \frac{\lambda}{N - 1} \). The sum of all weights $\sum\limits_{i = 1}^N {{w_i}}$ is constrained to 1. 

\textbf{Total objectives.} Finally, the total target training function can be summarized as the summation of the aforementioned three components:
\begin{equation}
\label{E_11}
{{\cal L}_{all}} = {{\cal L}_p} + {{\cal L}_b} + {{\cal L}_u}
\end{equation}

\section{Experiments \label{S_5}}
\subsection{Experimental Settings}
\subsubsection{Curated Datasets} MRG-MS-CXR \& MRG-ChestX-ray8 dataset. Quantitative assessments and benchmarks are crucial for grounding tasks in inferential medical reporting. To ensure reliable evaluation, we curated a benchmark dataset by collecting corresponding original reports from the MIMIC-CXR dataset~\cite{johnson2019mimic}. It comprises 1153 Image-Phrase-BBox triplet samples. To ensure that each sample consists of an X-ray image, medical report, phrase query, and bounding box quadruple, we performed preprocessing on the MS-CXR dataset. We named it as MRG-MS-CXR data, which encompassed diverse X-ray report diagnostic scenarios, as depicted in Fig.~\ref{F_1} \textbf{b}. Its diagnostic phrases mainly include cardiomegaly, pneumothorax, pleural effusion, etc. The MRG-MS-CXR benchmark comprises a total of 835 pairs of reports, images, grounding boxes, and phrases. Similarly, we constructed pairs from the ChestX-ray8 dataset \cite{wang2017chestx}. As the original ChestX-ray8 dataset only contains paired images, grounding boxes, and phrases, we used GPT-4 to expand each phrase into a corresponding report. This newly constructed paired dataset is referred to as MRG-ChestX-ray8. The MRG-ChestX-ray8 benchmark comprises a total of 984 paired reports, images, grounding boxes, and phrases. Following the~\cite{chen2023}, we split the two benchmarks into train-validation-test sets in a ratio of 7:1:2.

\subsubsection{Clinical Application Datasets} MRG-MIMIC-VQA \& MRG-MIMIC-Class dataset. To apply the proposed uMedGround to grounded medical visual question-and-answer (VQA) and class-based localization, we constructed sub-datasets based on the existing Medical-Diff-VQA~\cite{hu2023expert} and MIMIC-CXR datasets. For medical VQA, we extracted all VQA instances corresponding to MRG-MS-CXR from the Medical-Diff-VQA dataset, treating them as text inputs for uMedGround. This subset of the test dataset is named MRG-MIMIC-VQA. Similarly, we extracted the categories most relevant to MRG-MS-CXR from the MIMIC-CXR dataset and used GPT-4's descriptions of these categories as text inputs for uMedGround. We excluded samples not present in both datasets, resulting in a total of 158 paired VQA samples and 163 class-based localization samples corresponding to the MRG-MS-CXR test set. This subset of the test dataset is referred to as MRG-MIMIC-Class.

\subsubsection{Evaluation Metrics} To evaluate the effectiveness of the MRG task, we employ two types of evaluation metrics to measure the accuracy of phrase prediction and bounding box estimation. We adopt three widely used metrics from prior natural language generation (NLG) studies to evaluate the quality of predicted phrases. Specifically, BLEU measures n-gram precision between the generated and reference phrases; ROUGE-L captures recall-oriented similarity based on the longest common subsequence; and CIDEr emphasizes the relevance of informative phrases by incorporating term frequency-inverse document frequency weighting. To further evaluate the clinical and semantic fidelity of the predictions, we additionally incorporate three clinically meaningful metrics. CheXbert score~\cite{devlin2019bert} quantifies agreement with a CheXbert classifier across 14 clinical conditions, providing insight into diagnostic consistency. RadGraph F1~\cite{jain2021radgraph} assesses the overlap of predicted and ground-truth medical entities and relations within a structured radiology knowledge graph. Finally, BERTScore~\cite{smit2020chexbert} leverages contextual embeddings to capture deeper semantic similarity beyond surface-level lexical matching. Additionally, we utilize the widely-used metric mIoU (mean Intersection over Union) in bounding box prediction for a comprehensive comparison. Furthermore, we introduce the accuracy of bounding box prediction as an evaluation metric. Specifically, we consider a predicted bounding box as a positive sample if its mIoU with the ground truth bounding box exceeds certain thresholds (0.1, 0.3, and 0.5). We represent these thresholds as AP10 (mIOU $\textgreater$0.1), AP30 (mIOU $\textgreater$0.3), and AP50 (mIOU $\textgreater$0.5), respectively.

\subsubsection{Baselines} We compare our proposed method with several general and medical visual grounding approaches, including RefTR~\cite{RefTR2021}, VGTR~\cite{vgtr2022}, TransVG~\cite{deng2021transvg}, and MedRPG~\cite{chen2023}. All methods are evaluated under the same settings as our model to ensure a fair comparison. Since these baselines require pre-extracted medical phrases as input, we use GPT-4~\cite{nori2023capabilities}, LLaMA2~\cite{touvron2023llama}, and a fine-tuned version with LoRA adapter of LLaMA2-FT to extract phrases from medical reports during inference. In addition, we include comparisons with commonly used large language and vision-language models, such as GPT-4o~\cite{achiam2023gpt}, LLaVA-13B~\cite{liu2024llava}, and InternVL (specifically, InternVL2-8B)~\cite{chen2024internvl,chen2024far}. Notably, both LLaVA-13B and InternVL were further fine-tuned using LoRA adapter on our task-specific dataset and are referred to as LLaVA-FT and InternVL-FT, respectively.For all baseline methods, we utilize their official implementations and ensure that training and evaluation are conducted under consistent experimental conditions.

\subsubsection{Implementation Details} We conducted the following experiments on the PyTorch platform with an NVIDIA 48G A6000. The training scripts are based on deepspeed engine. We trained our model by the AdamW optimizer with the learning rate of 0.0003. Besides, the WarmupDecayLR strategy was employed for the learning rate scheduler and the its iterations are set to 100. The gradient accumulation step and batch size is set to 10 and 8, respectively. During training, the best checkpoints on the validation dataset of baselines and our model were reported with different random seeds.

\subsection{Experimental Results on MRG-MS-CXR dataset \label{S5_2}}
First, we conduct experiments on the MRG-MS-CXR dataset derived from MIMIC-CXR dataset~\cite{johnson2019mimic}, primarily demonstrating phrase identification, phrase grounding prediction, and visual comparison results.
\subsubsection{Phrase identification} It is worth noting that existing visual grounding approaches are not well-suited for the end-to-end medical report grounding task, as they are not designed to identify diagnostic phrases, which presents a unique and critical challenge in the clinical domain. To address this issue, we explored a range of language models to extract diagnostic phrases directly from medical reports, including GPT-4~\cite{achiam2023gpt} and LLaMA2~\cite{touvron2023llama} under a zero-shot setting. Additionally, we fine-tuned LLaMA2 using LoRA (referred to as LLaMA2-FT) and evaluated InternVL2-8B~\cite{chen2024internvl} on the same dataset to assess their capability in diagnostic phrase identification.
Tab.~\ref{T_1} summarizes the diagnostic phrase extraction performance of GPT-4, LLaMA2, LLaMA2-FT, InternVL2-8B, and our proposed method. In the zero-shot setting, GPT-4 performs poorly in identifying key phrases, while LLaMA2 shows moderate improvement but still falls short of clinical requirements. Although fine-tuning with LoRA allows LLaMA2-FT to surpass InternVL2-8B in overall performance, both models remain constrained in their ability to accurately capture domain-specific diagnostic content. In contrast, our proposed approach, uMedGround, demonstrates significantly superior performance in diagnostic phrase recognition. It achieves substantial gains across standard NLP metrics (BLEU, ROUGE-L, and CIDEr) as well as three clinically relevant evaluation metrics (CheXbert score, RadGraph F1 and BERTScore). These improvements are primarily attributed to our tailored fine-tuning of LoRA layers within a multimodal LLM, which is specifically optimized for foundational medical report understanding tasks. By fine-tuning LLaVA, we equip the model with effective diagnostic phrase recognition capabilities and transform it into a fully end-to-end multimodal LLM framework. This design eliminates the need for the traditional two-stage grounding box prediction pipeline, simplifying and enhancing the grounding process.

\begin{table*}[htbp]
  \centering
  \caption{Diagnostic phrase identification results on MRG-MS-CXR dataset with respect to BLEU\_1 , BLEU\_2, BLEU\_3, BLEU\_4, ROUGE\_L, CIDEr, CheXbert score, RadGraph F1 and BERTScore. Ours means the proposed method uMedGround.}
     \begin{tabular}{cccccccccc}
    \toprule
    Method  & BLEU\_1 & BLEU\_2 & BLEU\_3 & BLEU\_4 & ROUGE\_L & CIDEr & CheXbert score & RadGraph F1 & BERTScore \\
    \midrule
    GPT-4 & 14.81  & 11.34  & 8.95  & 6.69  & 17.76  & 0.60 & 0.5560  & 0.1464  & 0.3336   \\
    LLaMA2  & 15.89  & 12.58  & 10.25  & 8.49  & 20.06  & 0.88 & 0.5619  & 0.1799  & 0.3793  \\
    LLaMA2-FT  & 36.68 & 31.95 & 28.75 & 26.60 & 41.48 & 2.65 & 0.7571 & 0.3145 & 0.5151 \\
    InternVL2-8B  &33.34  & 28.11 & 24.49  & 21.79  & 20.06  & 2.06 & 0.7363  & 0.2558 & 0.4412  \\
    Ours  & \textbf{82.79}  & \textbf{82.11}  & \textbf{79.04}  & \textbf{77.80} & \textbf{65.90}  & \textbf{5.16} & \textbf{0.9224}  & \textbf{0.5590}   & \textbf{0.6990}  \\
    \bottomrule
    \end{tabular}
\label{T_1}
\end{table*}

\subsubsection{Phrase Grounding} Table~\ref{T_2} presents the comparison of grounding box prediction performance on the MRG-MS-CXR dataset. Since most baseline methods do not support end-to-end box prediction, we first employ GPT-4 and LLaMA2 in a zero-shot setting to extract diagnostic phrases that may facilitate visual grounding. We compare the results obtained by applying MedRPG~\cite{chen2023}, TransVG~\cite{deng2021transvg}, RefTR~\cite{RefTR2021}, and VGTR~\cite{vgtr2022} using these phrases as input. Among them, TransVG demonstrates better performance than RefTR and VGTR, with MedRPG closely following. However, all of these methods achieve relatively low mIoU scores, which may be attributed to the limited quality and specificity of the phrases extracted from diagnostic reports. To further explore this, we apply a fine-tuned version of LLaMA2 with LoRA adapter to extract more accurate diagnostic phrases under the same dataset settings. As shown in the third major block of Tab.~\ref{T_2}, this significantly improves the performance of both general-purpose and medical visual grounding baselines. Notably, our proposed method, MedGround, consistently outperforms other state-of-the-art approaches. In particular, it achieves an AP30 above 70\% and an AP10 close to 90\%. Compared with TransVG using LLaMA2-FT for phrase input, our MedGround and uMedGround models achieve the highest prediction performance. The uMedGround variant further enhances MedGround by incorporating uncertainty-aware prediction, demonstrating its effectiveness in improving grounding accuracy in medical scenarios.

In addition, we compare our method with several recent VLMs, including GPT-4o~\cite{achiam2023gpt}, LLaVA-13B~\cite{liu2024llava}, and InternVL~\cite{chen2024internvl} (specifically InternVL2-8B). To ensure a fair comparison, both LLaVA-13B and InternVL2-8B were fine-tuned using the LoRA adapter under same dataset setting as our uMedGround model. These fine-tuned models are referred to as LLaVA-FT and InternVL-FT, respectively. As shown in Tab.~\ref{T_2}, both LLaVA-FT and InternVL-FT achieve notable improvements over the original GPT-4o, with InternVL-FT exhibiting the strongest overall performance among the baseline VLMs. Nevertheless, our proposed MedGround and uMedGround still outperform InternVL-FT by a significant margin. Specifically, MedGround and uMedGround surpass InternVL-FT by over 30\% in AP50 and more than 20\% in mIoU. The performance gains are mainly driven by the use of a multimodal large language model, which extracts richer diagnostic phrases and enables the fine-tuned visual decoder to jointly learn the $<\mathtt{BOX}>$ token and visual features, leading to more consistent and precise grounding.
\begin{table}[h]
  \centering
  \renewcommand{\arraystretch}{1.4}
  \caption{Grounding box prediction on MRG-MS-CXR dataset with respect to AP10 (ACC(mIOU$\textgreater$0.1)), AP30 (ACC(mIOU$\textgreater$0.3)), and AP50 (ACC(mIOU$\textgreater$0.5)) and mIoU. ZS and FT mean the zero shot and fine-tuned, respectively. FT denotes the fine-tuned model. }
  \scalebox{1.1}{
    \begin{tabular}{cccccc}
    \toprule
    \multicolumn{2}{c}{Method} & AP10  & AP30  & AP50  & mIOU \\
    \midrule
    \multirow{4}[2]{*}{\begin{sideways}GPT-4\end{sideways}} & MedRPG~\cite{chen2023} & 57.49  & 40.12  & 18.56  & 25.73  \\
          & TransVG~\cite{deng2021transvg} & 59.28  & 40.72  & 22.16  & 26.28  \\
          & RefTR~\cite{RefTR2021} & 56.29  & 32.34  & 16.17  & 23.22  \\
          & VGTR~\cite{vgtr2022}  & 50.90  & 29.94  & 15.57  & 20.55  \\
    \midrule
    \midrule
    \multirow{4}[2]{*}{\begin{sideways}LLaMA2\end{sideways}} & MedRPG~\cite{chen2023} & 49.70  & 35.93  & 19.76  & 22.13  \\
          & TransVG~\cite{deng2021transvg} & 52.10  & 34.13  & 16.77  & 22.31  \\
          & RefTR~\cite{RefTR2021} & 55.09  & 32.34  & 16.17  & 21.75  \\
          & VGTR~\cite{vgtr2022}  & 47.31  & 28.74  & 14.37  & 19.11  \\
    \midrule
    \midrule
    \multirow{4}[2]{*}{\begin{sideways}LLaMA2-FT\end{sideways}} & MedRPG~\cite{chen2023} & 68.86  & 57.49  & 46.71  & 41.44  \\
          & TransVG~\cite{deng2021transvg} & 71.86  & 58.68  & 46.71  & 42.60  \\
          & RefTR~\cite{RefTR2021} & 69.46  & 53.29  & 43.11  & 38.19  \\
          & VGTR~\cite{vgtr2022}  & 72.46  & 51.50  & 41.92  & 38.06  \\
    \midrule
    \midrule
    \multicolumn{1}{c}{\multirow{5}[2]{*}{\begin{sideways}End-to-end\end{sideways}}}
    & GPT-4o~\cite{achiam2023gpt} & 24.55  & 8.98  & 2.99  & 7.51  \\
    & LLaVA-FT~\cite{liu2024llava} & 61.08  & 29.34  & 4.79  & 19.36  \\
    & InternVL-FT~\cite{chen2024internvl} & 57.49  & 36.53  & 17.96  & 23.21  \\
    & MedGround & 87.43  & 71.26  & 51.50  & 46.04  \\
    & uMedGround & \textbf{89.82}  & \textbf{72.46}  & \textbf{53.29}  & \textbf{47.65}  \\
    \bottomrule
    \end{tabular}}
  \label{T_2}%
\end{table}%

\begin{figure*}
\centering
\includegraphics[width=1\linewidth]{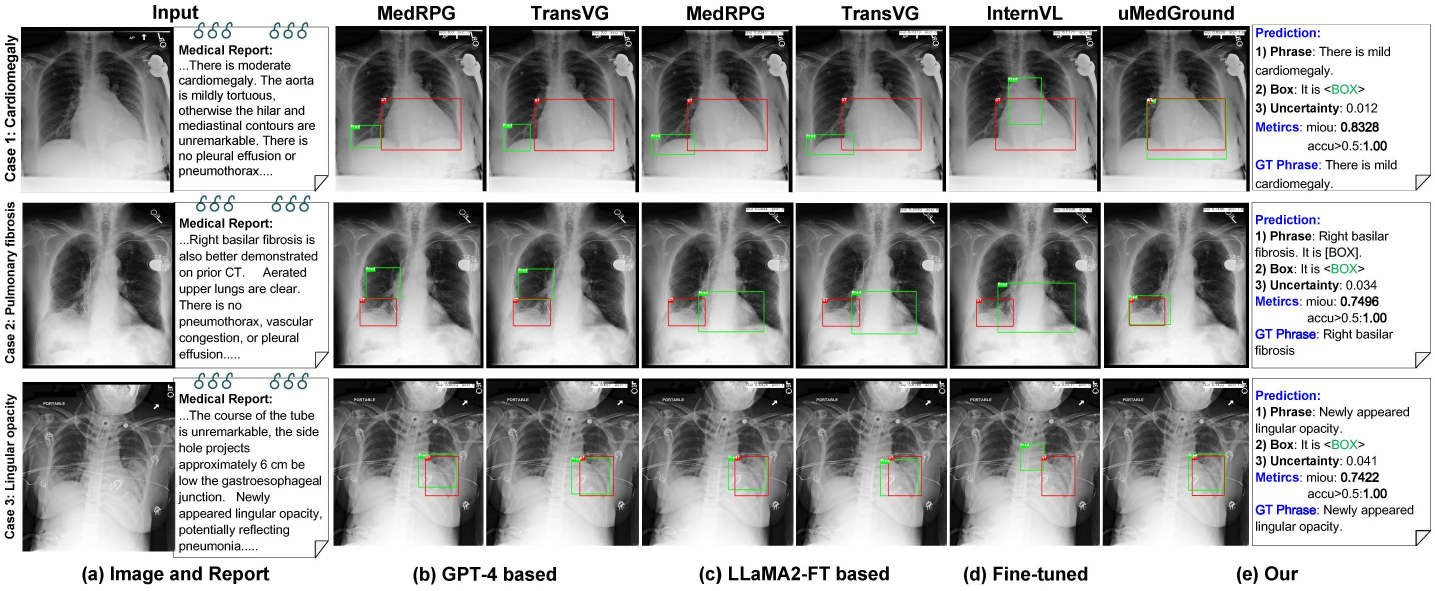}
\caption{Visual comparison of medical report grounding results across different methods on the MRG-ChestX-ray8 dataset. (a) Medical report and X-ray image input. (b) GPT-4 based methods for medical report grounding. (c) LLaMA2-FT based methods fine-tuned with LoRA for medical report grounding. (d) Fine-tuned method InterVL for medical report grounding. (e) Our proposed uMedGround method for medical report grounding.}
\label{F_3}
\end{figure*}

\subsubsection{Visual Comparisons} Figure~\ref{F_3} presents a visual comparison of three representative cases from the MRG-MS-CXR dataset, highlighting the grounding performance of our proposed method versus existing approaches. The focus is on evaluating the ability to localize key diagnostic phrases from medical reports. In Figures~\ref{F_3}\textbf{b)} and \textbf{c)}, we show the results of representative existing methods guided by different LLMs. Figures~\ref{F_3}\textbf{d)} and \textbf{e)} illustrate the performance of the fine-tuned InternVL model and our proposed uMedGround approach, respectively. In Case 1 (cardiomegaly), existing methods such as MedRPG~\cite{chen2023} and TransVG~\cite{deng2021transvg} demonstrate poor grounding performance when guided by phrases extracted using GPT-4. When guided by phrases extracted via a LLaMA2 model fine-tuned with LoRA adapters, these methods show slightly improved localization performance, although the mIoU remains close to 10\%. The fine-tuned InternVL model offers little improvement but still exhibits missed detections or false positives, achieving a much lower mIoU compared to our uMedGround, which reaches 83.28\%. A similar trend is observed in Case 2. While MedRPG and TransVG again perform slightly better under the guidance of LLaMA2-FT compared to GPT-4, their performance remains suboptimal. The fine-tuned InternVL also underperforms, whereas uMedGround achieves an impressive mIoU of 74.96\%. In Case 3, where the grounding task is relatively easier, all methods show improved localization. MedRPG and TransVG, guided by phrases from LoRA-tuned LLaMA2, outperform their GPT-4 counterparts. All four methods achieve approximately 70\% mIoU, which is significantly better than the fine-tuned InternVL. Notably, MedRPG and TransVG generally outperform InternVL under LoRA-tuned LLaMA2 guidance due to their designs being more tightly coupled with visual-language grounding tasks. Nonetheless, our uMedGround approach still achieves the highest accuracy, with mIoU close to 75\%. Overall, uMedGround not only delivers more accurate grounding box predictions but also reliably identifies diagnostic phrases. This effectiveness is largely attributed to the use of large language models fine-tuned with LoRA adapters for diagnostic phrase extraction, and their integration into the visual decoder for grounding. These findings demonstrate the feasibility and strong performance of our proposed medical report grounding framework, setting a new state-of-the-art and supporting the broader application of large language models in medical imaging tasks.

\subsection{Experimental Results on MRG-ChestX-ray8 dataset \label{S5_b2}}
Furthermore, we conduct experiments on the MRG-ChestX-ray8 dataset derived from Chest-Xray8. Unlike the MRG-MS-CXR dataset, this dataset lacks original medical reports and primarily consists of simulated reports generated by GPT-4. Therefore, this section focuses on demonstrating phrase grounding and visual comparison results.

\subsubsection{Phrase Grounding} Tab.~\ref{T_3} presents the comparison results of phrase grounding prediction on the MRG-ChestX-ray8 dataset. To adapt the ChestX-ray8 dataset to the medical report grounding task, we first utilized GPT-4 and LLaMA2 in a zero-shot setting to expand the original short phrases into medical-report-style inputs. We then evaluated several existing grounding methods, MedRPG~\cite{chen2023}, TransVG~\cite{deng2021transvg}, RefTR~\cite{RefTR2021}, and VGTR~\cite{vgtr2022}, with the expanded medical reports as input. As shown in the first two major blocks of Tab.~\ref{T_3}, MedRPG consistently outperforms these methods under both GPT-4 and LLaMA2 expansions. To enable a fairer comparison, we further evaluated these methods using the original phrase annotations from MRG-ChestX-ray8 dataset, without any expansion into report-style inputs. The corresponding results are reported in the third major block of Table~\ref{T_3}, where we observe a notable improvement in phrase grounding accuracy. Despite these gains, our proposed models, MedGround and uMedGround, still outperform all existing baselines, particularly in AP metrics ranging from AP10 to AP50.

We also benchmarked end-to-end VLMs including GPT-4o, LLaVA-13B, and InternVL2-8B. All models received the same GPT-4 expanded medical reports as input to ensure consistency. GPT-4o was evaluated under a zero-shot setting, whereas LLaVA-13B and InternVL2-8B were fine-tuned using LoRA-based adapters under the same dataset setting as our method, and are thus denoted as LLaVA-FT and InternVL-FT, respectively. Among these VLMs, InternVL-FT achieved the best overall performance, reaching over 70\% AP10, surpassing MedRPG under phrase-based input, but still underperformed in other metrics compared to MedRPG. Despite the improvements from recent VLMs, MedGround and uMedGround consistently achieved superior performance across all evaluation metrics. Notably, uMedGround outperformed the best-performing baseline by more than 10\% in AP10 and 4\% in mIoU. These improvements stem from our framework’s capacity to extract key diagnostic phrases through a multimodal language model and integrate them into the visual decoder, which enhances grounding box consistency and relevance.

\begin{figure*}
\centering
\includegraphics[width=1\linewidth]{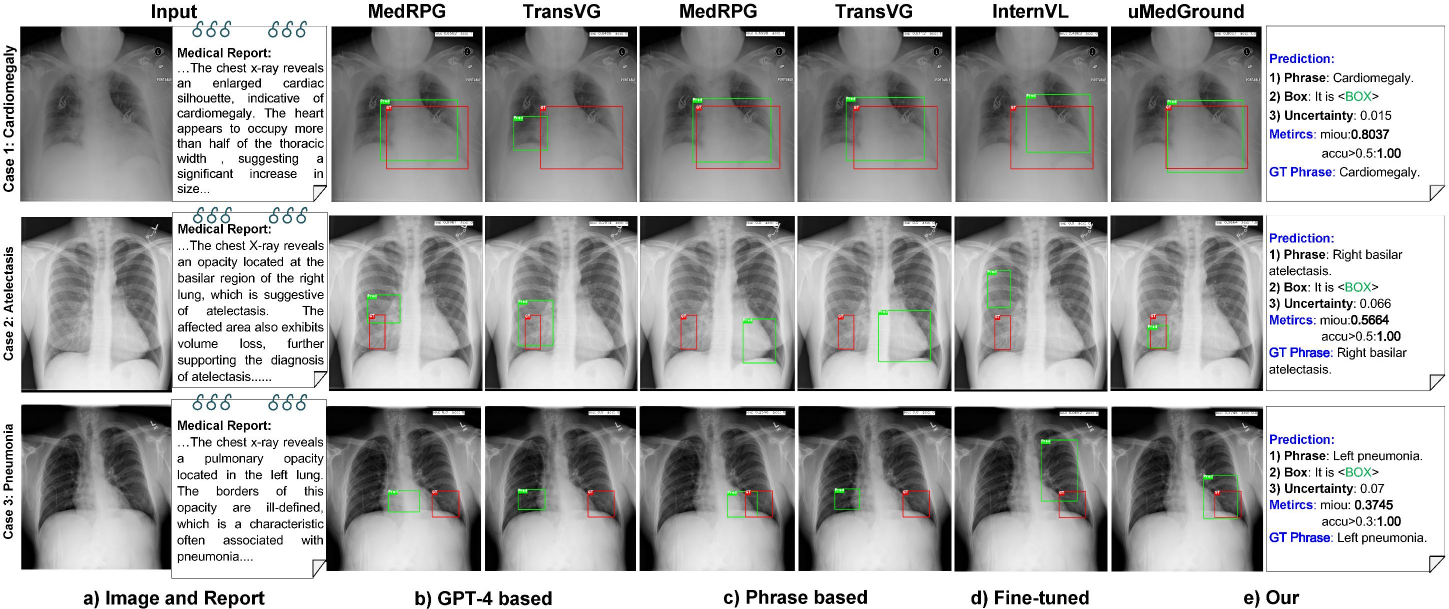}
\caption{Medical report grounding results of our method and competing baselines on the differennt cases. (a) Input medical report and corresponding chest X-ray image. (b) Grounding results using GPT-4-expanded medical reports. (c) Grounding based on ground-truth diagnostic phrases. (d) Grounding results from LoRA-fine-tuned InternVL (denoted as InternVL-FT). (e) Results of our proposed uMedGround method. }
\label{F_3_1}
\end{figure*}

\begin{table}[htbp]
  \centering
  \caption{Grounding box prediction on MRG-ChestX-ray8 dataset with respect to AP10 (ACC(mIOU$\textgreater$0.1)), AP30 (ACC(mIOU$\textgreater$0.3)), and AP50 (ACC(mIOU$\textgreater$0.5)) and mIoU. FT denotes the fine-tuned model.}
  \renewcommand{\arraystretch}{1.1}
  \scalebox{1.1}{
    \begin{tabular}{cccccc}
    \toprule
    \multicolumn{2}{c}{Method} & AP10  & AP30  & AP50  & mIOU \\
    \midrule
    \multicolumn{1}{c}{\multirow{4}[2]{*}{\begin{sideways}GPT-4\end{sideways}}} & MedRPG~\cite{chen2023} & 53.54  & 37.37 & 27.78 & 27.67  \\
    & TransVG~\cite{deng2021transvg} & 54.04  & 37.37  & 25.76  & 26.27\\
    & RefTR~\cite{RefTR2021} & 51.51  & 38.38  & 25.25  & 26.27\\  
    & VGTR~\cite{vgtr2022}  & 56.06  & 27.78  & 9.60  & 19.28  \\
    \midrule
    \midrule
    \multicolumn{1}{c}{\multirow{4}[2]{*}{\begin{sideways}LLaMA2\end{sideways}}} & MedRPG~\cite{chen2023} & 50.51  & 34.85  & 24.24  & 25.53  \\
    & TransVG~\cite{deng2021transvg} & 46.46 & 33.84  & 25.25  & 23.42  \\
    & RefTR~\cite{RefTR2021} & 58.08  & 33.33  & 13.64  & 21.38 \\
    & VGTR~\cite{vgtr2022}  & 56.57 & 23.23  & 7.07  & 17.65  \\
    \midrule
    \midrule
    \multicolumn{1}{c}{\multirow{4}[2]{*}{\begin{sideways}Phrase\end{sideways}}} & MedRPG~\cite{chen2023} & 65.66  & 48.48  & 32.83  & 34.17  \\
    & TransVG~\cite{deng2021transvg} & 63.13 & 45.96  & 31.31  & 32.48  \\
    & RefTR~\cite{RefTR2021} & 61.62  & 45.45  & 30.81  & 30.98\\
    & VGTR~\cite{vgtr2022}  & 63.13  & 46.97  & 29.29  & 29.9  \\
    \midrule
    \midrule
    \multicolumn{1}{c}{\multirow{5}[2]{*}{\begin{sideways}End-to-end\end{sideways}}}
    & GPT-4o~\cite{achiam2023gpt} & 25.76  & 18.69  & 6.06  & 10.12  \\
    & LLaVA-FT~\cite{liu2024llava} & 46.97  & 24.24  & 11.62  & 17.63  \\
    & InternVL-FT~\cite{chen2024internvl} & 70.71  & 44.44  & 20.71  & 27.93  \\
    & MedGround   & 82.32  & 55.05  & 31.82  & 35.91  \\
          & uMedGround  & \textbf{84.34}  & \textbf{56.57}  & \textbf{38.38}  & \textbf{38.49}  \\
    \bottomrule
    \end{tabular}%
    }
  \label{T_3}%
\end{table}%

\subsubsection{Visual Comparisons} Fig.~\ref{F_3_1} presents a visual comparison of representative methods on three cases from the MRG-ChestX-ray8 dataset. Fig.~\ref{F_3_1} \textbf{b)} and \textbf{c)} illustrate the visual grounding results of MedRPG and TransVG, using either GPT-4-augmented reports or ground-truth phrases as input. Both methods generally exhibit suboptimal performance, often missing critical regions or producing false positives. Similar issues are observed in InternVL, a fine-tuned VLM on this task, as shown in the Fig.~\ref{F_3_1} \textbf{c)}. In the simpler Case 1 (cardiomegaly), TransVG, MedRPG, and InternVL methods approximately localize the correct region, with MedRPG demonstrating superior phrase grounding when the original phrase is provided as input. In contrast, uMedGround achieves the closest alignment with the ground truth. In the more challenging Case 2 (Atelectasis), InternVL struggles to localize the target phrase, while MedRPG and TransVG with GPT-generated inputs succeed in identifying relevant regions but introduce false positive detections. When using ground-truth phrases as input, MedRPG and TransVG incorrectly focus on the cardiac area, likely due to data imbalance favoring that region. Despite missing certain regions, uMedGround identifies phrase-relevant areas that largely overlap with the annotated targets. Finally, in Case 3 (Pneumonia), TransVG performs poorly under both GPT-augmented and phrase-based inputs, while MedRPG shows modest improvement with phrase input but still underperforms compared to uMedGround. Overall, across diverse clinical scenarios, uMedGround consistently outperforms prior methods by more precisely aligning visual evidence with textual descriptions, resulting in improved visual grounding performance. 

\subsection{Performance of uMedGround on simulation Data} 
To evaluate the robustness of different grounding methods under low-quality image conditions, we introduced Gaussian noise into the original X-ray images. As shown in Table~\ref{T_five}, we compared our proposed method with existing approaches under varying levels of Gaussian noise (${\sigma^2} = 0.01$ and ${\sigma^2} = 0.1$). The results demonstrate that our uMedGround method exhibits superior robustness to noise. Specifically, on the MRG-MS-CXR dataset, under the guidance of GPT-4 extracted phrases, TransVG experienced a performance drop of 5.99\% in AP30 and 3.76\% in mIoU. MedRPG showed a similar level of degradation. When guided by phrases extracted from a LLaMA2 model fine-tuned with LoRA adapters, the performance drop of TransVG slightly improved to 4.19\% (AP30) and 2.32\% (mIoU), with MedRPG also exhibiting a moderate decline. In contrast, our uMedGround method showed only a 0.64\% decrease in mIoU under the same noise conditions. On the MRG-ChestX-ray8 dataset, we applied a stronger level of Gaussian noise to further assess model stability. Both MedRPG and TransVG exhibited noticeable performance degradation under noisy conditions, regardless of whether the inputs were GPT-4 generated reports or manually extracted diagnostic phrases. In comparison, our uMedGround approach maintained stable performance, with only a 1.49\% reduction in mIoU under high noise levels. Overall, the uncertainty estimation mechanism introduced in our medical report grounding framework, UPM, enables confidence prediction for the generated outputs and simultaneously enhances the model’s robustness. This improvement can be attributed to UPM’s multi-hypothesis design, which generates multiple plausible predictions and preserves output diversity, rather than relying solely on a single deterministic output.

\begin{figure*}
\centering
\includegraphics[width=1\linewidth]{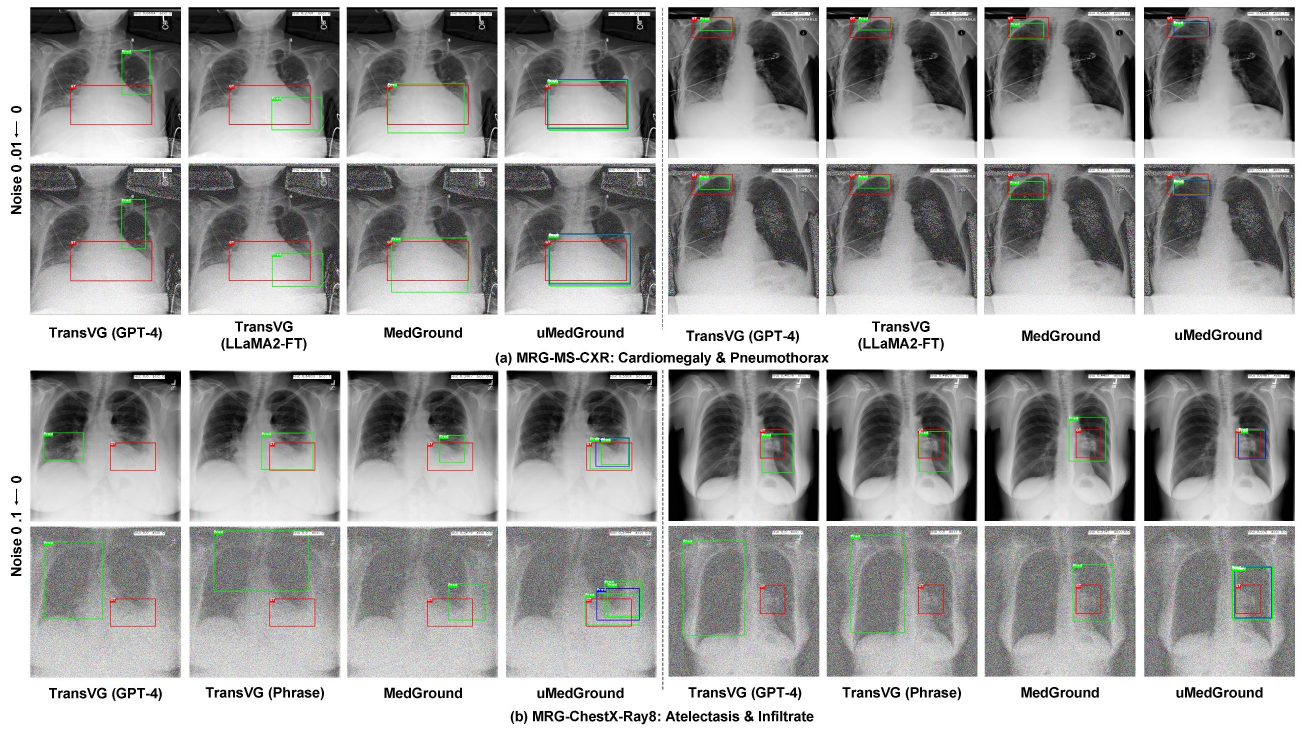}
\caption{(a–b) Grounding results under normal and Gaussian noise conditions on the MRG-MS-CXR and MRG-ChestX-ray8 datasets using TransVG (GPT-4), TransVG (LLaMA2-FT), TransVG (Phrase), MedGround, and uMedGround. In (a), TransVG (GPT-4) and TransVG (LLaMA2-FT) use diagnostic phrases extracted from medical reports by GPT-4 and LoRA-fine-tuned LLaMA2, respectively. In (b), TransVG (GPT-4) is based on GPT-4-expanded medical reports, while TransVG (Phrase) uses ground-truth diagnostic phrases extracted from the reports.}
\label{F_5}
\end{figure*}

In addition, we provide qualitative visualization results under different noise levels, as illustrated in Figure~\ref{F_5}. Across both datasets, the introduction of Gaussian noise (${\sigma^2} = 0.01$ and ${\sigma^2} = 0.1$) led to visibly degraded grounding box predictions in most baseline methods. Both TransVG and MedRPG were notably affected under various noise conditions. In contrast, our uMedGround model exhibited more stable and consistent grounding results, highlighting the advantage of incorporating uncertainty estimation.

\begin{table}[h]
  \centering
    \renewcommand{\arraystretch}{1.4}
  \caption{Grounding box prediction results of different methods on low-quality images: MRG-MS-CXR (Gaussian noise $\sigma^2=0.01$) and MRG-ChestX-ray8 (Gaussian noise $\sigma^2=0.1$).}
  \scalebox{1.15}{
    \begin{tabular}{cccccc}
    \toprule
    \multicolumn{2}{c}{\multirow{2}{*}{Method}} & \multicolumn{2}{c}{MRG-MS-CXR} & \multicolumn{2}{c}{MRG-ChestX-ray8} \\
    \multicolumn{2}{c}{} & AP30 & mIOU & AP30 & mIOU \\
    \midrule
    \multicolumn{1}{c}{\multirow{2}[2]{*}{\begin{sideways}GPT-4\end{sideways}}}
    & MedRPG~\cite{chen2023} & 27.54 & 19.92 & 9.09  & 7.84 \\
    & TransVG~\cite{deng2021transvg} & 34.73  & 22.52  & 25.76 & 16.69 \\
    \midrule
    \midrule
    \multicolumn{1}{c}{\multirow{2}[2]{*}{\begin{sideways}FT/P\end{sideways}}}
    & MedRPG~\cite{chen2023} & 53.89 & 39.82 & 10.61  & 9.69 \\
    & TransVG~\cite{deng2021transvg} & 54.49  & 40.28  & 25.76 & 18.74 \\
    \midrule
    \midrule
    \multicolumn{1}{c}{\multirow{2}[2]{*}{\begin{sideways}Ours\end{sideways}}}
    & MedGround & 70.06  & 44.65  & 48.48 & 32.52 \\
    & uMedGround & \textbf{72.46}  & \textbf{47.01}  & \textbf{55.05}  & \textbf{35.16} \\
    \bottomrule
    \end{tabular}
  }
  \label{T_five}
\end{table}

\subsection{Generalization Performance of uMedGround} 
To validate the generalization capability of the proposed model, we conducted a cross-dataset test. We used the model trained on the MRG-MS-CXR dataset to test the MRG-ChestX-ray8 dataset. Tab.~\ref{T_4} presents the results of different methods in terms of generalization performance. Although our proposed method shows lower performance in the AP50 metric compared to existing methods such as MedRPG and TransVG, it outperforms them in other metrics, including mIOU. Notably, the AP10 metric is 9.09\% higher than the best existing method. This demonstrates that our model can promptly predict grounding boxes on other datasets with lower IOU thresholds, which is crucial for alerting clinical staff to abnormal conditions.

\begin{table}[htbp]
  \centering
    \renewcommand{\arraystretch}{1.4}
  \caption{Generalization of grounding box prediction on MRG-ChestX-ray8 dataset (Training on MRG-MS-CXR dataset) with respect to AP10 (ACC(mIOU$\textgreater$0.1)), AP30 (ACC(mIOU$\textgreater$0.3)), and AP50 (ACC(mIOU$\textgreater$0.5)) and mIoU.}
  \scalebox{1.15}{
    \begin{tabular}{ccccc}
    \toprule
    Method & AP10  & AP30  & AP50  & mIOU \\
    \midrule
    MedRPG~\cite{chen2023} & 66.16  & 40.40  & 26.26  & 28.35  \\
    TransVG~\cite{deng2021transvg} & 62.63  & 39.39  & 23.74  & 26.63  \\
    \midrule
    \midrule
     MedGround   & 73.23  & 40.91  & 21.72  & 28.38  \\
     uMedGround  & \textbf{75.25}  & \textbf{42.42} & \textbf{23.74}  & \textbf{30.43}  \\
    \bottomrule
    \end{tabular}%
    }
  \label{T_4}%
\end{table}%

\subsection{Clinical Extension of uMedGround} 
As shown in Fig.~\ref{F_2} {\bf{c}}, we aim to leverage the extension of uMedGround for broader potential clinical applications. To this end, we applied it to primary healthcare VQA tasks and class-based localization tasks. For medical VQA, we used QA pairs as the textual input to uMedGround, demonstrating that medical QA can effectively serve as input for our model. Furthermore, in more complex clinical scenarios, physicians may provide only a textual category rather than detailed diagnostic insights. Based on this observation, we used category-based text as the textual input to uMedGround to show that it can also function as effective input for our model. Consequently, we constructed two clinical extension application sub-test datasets, MRG-MIMIC-VQA and MRG-MIMIC-Class, for experimentation.

The experimental results are summarized in Table~\ref{T_application}. By observing the first and second rows, it is evident that the uMedGround model performs significantly better with VQA-based report descriptions generated by GPT-4 compared to class-based inputs. This improvement may be due to the fact that VQA-based descriptions are more specific than class-based inputs and may include location information. Furthermore, when compared to using the original report as input, the results show that the original report is better than the VQA-based description.

\begin{figure*}[!t]
\centering
\includegraphics[width=1\linewidth]{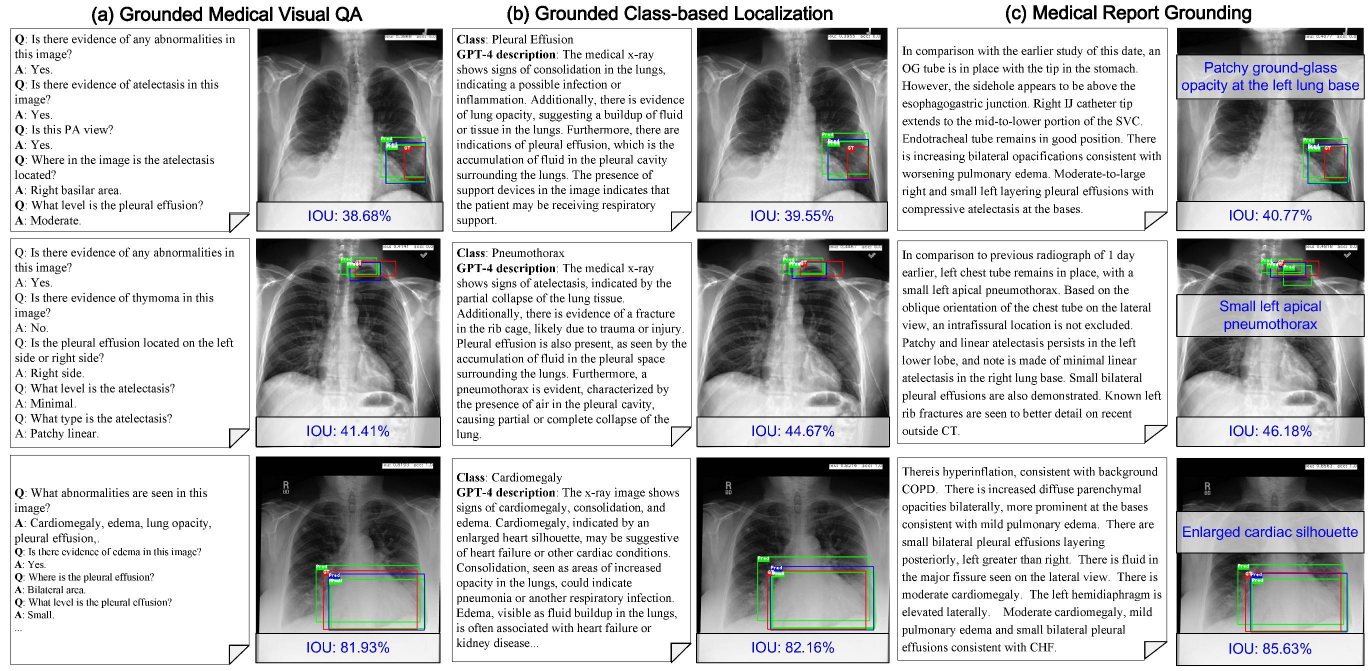}
\caption{Visual results on clinical applications. (a) Grounded medical visual question-and-answer (Medical question-and-answer as as text input). (b) Grounded Class-based localization (Class-based words with GPT-4 description as as text input). (c) Medical Report Grounding (Original report as text input). }
\label{F_application}
\end{figure*}

In addition, Figure~\ref{F_application} presents the visualization results of grounding box predictions under different cases with varying textual inputs. As shown in Figure~\ref{F_application}(a), uMedGround successfully predicts the phrases and their corresponding grounding boxes. Furthermore, when class information is integrated as the textual input to uMedGround, the model's predictions for phrases and grounding boxes are consistent and outperform those using QA-based inputs. Finally, when the original report is used as the textual input to uMedGround, the model achieves the best results for both phrase and grounding box predictions. The above experiments validate the extension of uMedGround in clinical applications, demonstrating that even when QA-based inputs are used, the proposed model may still achieve pre-satisfactory results.

\label{sec_s4}
\begin{table}[h]
  \centering
  \caption{Question-and-answer pairs, class descriptions generated by GPT-4, and original medical report as quantitative results for uMedGround.}
    \begin{tabular}{ccrrrr}
    \toprule
    \multicolumn{2}{c}{Text Input} & \multicolumn{1}{c}{AP10} & \multicolumn{1}{c}{AP30} & \multicolumn{1}{c}{AP50} & \multicolumn{1}{c}{mIOU} \\
    \midrule
    \multicolumn{2}{c}{MRG-MIMIC-Class} & 87.97 & \textbf{73.42} & 49.37 & 46.61 \\
    \multicolumn{2}{c}{MRG-MIMIC-VQA } & 88.47 & 73.39 & 50.55 & 47.39 \\
    \multicolumn{2}{c}{MRG-MS-CXR} & \textbf{89.82}  & 72.46  & \textbf{53.29}  & \textbf{47.65}   \\
    \bottomrule
    \end{tabular}%
  \label{T_application}%
\end{table}%

\subsection{Ablation Study and Analysis}
\label{sec_s1}
In this section, we present an ablation study to analyze the impact of different components within the proposed uMedGround framework. As shown in Tab.~\ref{T_5}, uMedGround consists of three core modules: LLaVA, SAM, and the uncertain prediction model.

A natural approach to predict grounding boxes is to leverage LLaVA by embedding a special token $<BOX>$ and directly regressing its coordinates. Due to GPU memory limitations, we adopt LoRA-based fine-tuning for LLaVA. As shown in the first row of Table~\ref{T_5}, this configuration yields limited performance in box prediction. This may be attributed to the fact that LoRA tuning mainly enhances the language component for medical phrase extraction, while contributing less to visual feature learning. To improve visual representation, we introduce the SAM~\cite{ma2024segment} as the visual backbone. We investigate three training strategies: training only the decoder while freezing the encoder, applying LoRA to the encoder, and training both the encoder and decoder. The result from training only the decoder is presented in the second row of Table~\ref{T_5}, which clearly outperforms the LLaVA-only setup. This demonstrates the strong capability of SAM, based on the ViT-H architecture, in extracting visual cues relevant to medical grounding. We further examine how these three SAM configurations interact with the uncertainty-aware prediction module, which produces multiple hypotheses to improve prediction reliability. The results are shown in the third to fifth rows of Table~\ref{T_5}. Among these, the best performance is observed when the SAM encoder is kept frozen and only the decoder is trained. This suggests that fine-tuning the encoder in the medical domain may lead to overfitting, especially when both encoder and decoder are updated simultaneously. At last, the best-performing configuration, which we refer to as uMedGround, combines LoRA-tuned LLaVA, SAM with a frozen encoder and a trained decoder, and the uncertainty-aware prediction module. This combination effectively integrates multimodal language understanding with robust visual grounding, leading to improved accuracy and robustness in medical report grounding tasks.

\begin{table}[h]
  \centering
  \caption{Ablation studies examining the impact of different modules on grounding box predictions. \checkmark indicates the module is included. Numbers in \textbf{bold font} indicate the best results. 
  $D_{FT}$ denotes fine-tuning only the decoder; $E_{lr}$+$D_{FT}$ denotes fine-tuning the encoder with the LoRA adapter together with full decoder fine-tuning; $E_{FT}$+$D_{FT}$ denotes full fine-tuning of both the encoder and decoder.}
  \scalebox{0.85}{
  \begin{tabular}{c ccc c cc}
    \toprule
    \multirow{2}{*}{\centering LLaVA+LoRA} & \multicolumn{3}{c}{SAM} & \multirow{2}{*}{\centering UPM} & \multirow{2}{*}{\centering AP30} & \multirow{2}{*}{\centering mIOU} \\
    \cmidrule(lr){2-4}
    & $D_{FT}$ & $E_{lr}$+$D_{FT}$ & $E_{FT}$+$D_{FT}$ & & & \\
    \midrule
    \checkmark &       &       &       &       & 29.34 & 19.36 \\
    \checkmark & \checkmark &       &       &       & 71.26 & 46.04 \\
    \checkmark & \checkmark &       &       & \checkmark & \textbf{72.46} & \textbf{47.65} \\
    \checkmark &       & \checkmark &       & \checkmark & 67.07 & 45.31 \\
    \checkmark &       &       & \checkmark & \checkmark & 66.10 & 41.53 \\
    \bottomrule
  \end{tabular}
  }
  \label{T_5}
\end{table}

\section{Discussion}
The proposed uMedGround framework demonstrates consistent and significant improvements across multiple datasets and evaluation dimensions. On the MRG-MS-CXR benchmark, our model outperforms various baselines not only on standard NLG metrics such as BLEU, ROUGE-L, and CIDEr, but also on clinically meaningful metrics including CheXbert score, RadGraph F1, and BERTScore. For the phrase grounding task, uMedGround surpasses both general visual grounding and specialized medical phrase grounding approaches. Even when using diagnostic phrases extracted by fine-tuned LLaMA, our model consistently achieves better performance. Moreover, uMedGround outperforms leading end-to-end vision-language models such as GPT-4o, LLaVA-13B, and fine-tuned InternVL, highlighting its ability to generate semantically accurate and clinically aligned diagnostic phrases and grounding outputs.

Similar findings hold on the MRG-ChestX-ray8 dataset, where uMedGround continues to deliver superior performance, especially when diagnostic phrases are used as input to other non end-to-end grounding methods. This consistency across datasets underscores the framework’s robust capabilities in both differential diagnostic phrase identification and visual grounding.

Qualitative visualizations further illustrate that uMedGround produces more precise and semantically meaningful bounding boxes than existing medical phrase grounding methods. The multi-hypothesis grounding module improves reliability by modeling local uncertainty, which is crucial for building clinician trust. This feature is especially valuable in complex or low-quality chest X-rays, where deterministic models may produce overconfident but incorrect predictions. Experiments on synthetic data confirm that uMedGround maintains robustness under image degradation, ensuring consistent grounding performance. Additionally, cross-validation across two benchmarks demonstrates strong transferability with minimal performance degradation, further supporting its potential for clinical deployment.

We also extended uMedGround to real-world clinical tasks such as primary care VQA and class-based localization. These experiments show that our model can flexibly handle various forms of textual input, including free-text reports, VQA queries, and class labels, and accurately identify the corresponding visual evidence (grounding box). This versatility is critical for supporting clinicians in interpreting diverse types of input within practical workflows.

Despite these encouraging results, several limitations remain. First, although the multi-hypothesis grounding module is designed to be lightweight, it introduces a slight computational overhead during inference, which may pose challenges in resource-constrained settings. Second, performance may decline when the model encounters rare or novel diagnostic phrases that lack well-defined spatial representations in the training data. Future work could address this by integrating external retrieval-augmented modules to expand semantic coverage. Moreover, while our current evaluation relies on automated metrics, it is necessary to validate the framework’s utility through prospective clinical studies, such as expert-in-the-loop evaluations in real-world settings.

In summary, uMedGround achieves a balanced integration of model performance, trustworthiness, and clinical utility. Its architecture, which features uncertainty-aware prediction and minimal reliance on handcrafted phrase annotations, makes it particularly well suited for deployment in real-world radiology workflows. The framework supports a range of tasks including diagnostic report generation, clinical question answering, and class-based phrase localization, demonstrating its versatility and potential to serve as a core component of multimodal medical AI systems in practical clinical settings.

\section{Conclusions}
In conclusion, we introduce the MRG task as a novel approach to enhance medical image analysis and radiological diagnosis. Our proposed uMedGround model combines a multimodal LLM with a specially designed $<$$\mathtt{BOX}$$>$ token for embedding-based detection and integrates uncertainty-aware prediction to boost grounding box confidence and reliability. Extensive experiments demonstrate that uMedGround consistently outperforms state-of-the-art visual grounding methods and vision-language models in terms of both accuracy and robustness. Furthermore, we investigate the clinical applicability of various text inputs for grounding, showcasing the adaptability of our approach to diverse medical scenarios.

This work represents a significant advancement in multimodal medical image analysis, addressing critical challenges in radiological diagnosis by providing precise and interpretable grounding results. By bridging the gap between textual medical reports and visual cues in medical images, our approach lays the groundwork for improved diagnostic workflows. In future work, we plan to extend our research in two directions. First, we will construct larger and more generalizable datasets to further evaluate the robustness of our method~\citep{liu2024gemex}. Second, we aim to integrate the system into real-world clinical workflows by collaborating with resident physicians to conduct human evaluations, exploring its potential for deployment in routine practice~\citep{wang2025enhancing}.

\bibliographystyle{IEEEtran}
\bibliography{Ucerntainty_review}

\begin{thebibliography}{10}
\providecommand{\url}[1]{#1}
\csname url@samestyle\endcsname
\providecommand{\newblock}{\relax}
\providecommand{\bibinfo}[2]{#2}
\providecommand{\BIBentrySTDinterwordspacing}{\spaceskip=0pt\relax}
\providecommand{\BIBentryALTinterwordstretchfactor}{4}
\providecommand{\BIBentryALTinterwordspacing}{\spaceskip=\fontdimen2\font plus
\BIBentryALTinterwordstretchfactor\fontdimen3\font minus \fontdimen4\font\relax}
\providecommand{\BIBforeignlanguage}[2]{{%
\expandafter\ifx\csname l@#1\endcsname\relax
\typeout{** WARNING: IEEEtran.bst: No hyphenation pattern has been}%
\typeout{** loaded for the language `#1'. Using the pattern for}%
\typeout{** the default language instead.}%
\else
\language=\csname l@#1\endcsname
\fi
#2}}
\providecommand{\BIBdecl}{\relax}
\BIBdecl

\bibitem{chen2023}
Z.~Chen, Y.~Zhou, A.~Tran, J.~Zhao, and et~al., ``Medical phrase grounding with region-phrase context contrastive alignment,'' in \emph{Medical Image Computing and Computer Assisted Intervention -- MICCAI 2023}.\hskip 1em plus 0.5em minus 0.4em\relax Cham: Springer Nature Switzerland, 2023, pp. 371--381.

\bibitem{deng2021transvg}
J.~Deng, Z.~Yang, T.~Chen, W.~Zhou, and H.~Li, ``Transvg: End-to-end visual grounding with transformers,'' in \emph{{ICCV}}, 2021, pp. 1769--1779.

\bibitem{vgtr2022}
Y.~Du, Z.~Fu, Q.~Liu, and Y.~Wang, ``Visual grounding with transformers,'' in \emph{{ICME}}.\hskip 1em plus 0.5em minus 0.4em\relax IEEE, 2022, pp. 1--6.

\bibitem{zhu2022seqtr}
C.~Zhu, Y.~Zhou, Y.~Shen, G.~Luo, X.~Pan, M.~Lin, C.~Chen, L.~Cao, X.~Sun, and R.~Ji, ``Seqtr: A simple yet universal network for visual grounding,'' in \emph{{ECCV}}.\hskip 1em plus 0.5em minus 0.4em\relax Springer, 2022, pp. 598--615.

\bibitem{zhou2023joint}
L.~Zhou, Z.~Zhou, K.~Mao, and Z.~He, ``Joint visual grounding and tracking with natural language specification,'' in \emph{{CVPR}}, 2023, pp. 23\,151--23\,160.

\bibitem{he2024improved}
R.~He, P.~Cascante-Bonilla, Z.~Yang, A.~C. Berg, and V.~Ordonez, ``Improved visual grounding through self-consistent explanations,'' in \emph{{CVPR}}, 2024, pp. 13\,095--13\,105.

\bibitem{Ichinose2023}
A.~Ichinose, T.~Hatsutani, K.~Nakamura, Y.~Kitamura, S.~Iizuka, E.~Simo-Serra, S.~Kido, and N.~Tomiyama, ``Visual grounding of whole radiology reports for 3d ct images,'' in \emph{{MICCAI}}.\hskip 1em plus 0.5em minus 0.4em\relax Springer, 2023, pp. 611--621.

\bibitem{achiam2023gpt}
J.~Achiam, S.~Adler, S.~Agarwal, L.~Ahmad, I.~Akkaya, F.~L. Aleman, D.~Almeida, J.~Altenschmidt, S.~Altman, S.~Anadkat \emph{et~al.}, ``Gpt-4 technical report,'' \emph{arXiv preprint arXiv:2303.08774}, 2023.

\bibitem{liu2024llava}
H.~Liu, C.~Li, Q.~Wu, and Y.~J. Lee, ``Visual instruction tuning,'' \emph{{NeurIPS}}, vol.~36, 2024.

\bibitem{chen2024internvl}
Z.~Chen, J.~Wu, W.~Wang, W.~Su, G.~Chen, S.~Xing, M.~Zhong, Q.~Zhang, X.~Zhu, L.~Lu \emph{et~al.}, ``Internvl: Scaling up vision foundation models and aligning for generic visual-linguistic tasks,'' in \emph{{CVPR}}, 2024, pp. 24\,185--24\,198.

\bibitem{chen2024far}
Z.~Chen, W.~Wang, H.~Tian, S.~Ye, Z.~Gao, E.~Cui, W.~Tong, K.~Hu, J.~Luo, Z.~Ma \emph{et~al.}, ``How far are we to gpt-4v? closing the gap to commercial multimodal models with open-source suites,'' \emph{arXiv preprint arXiv:2404.16821}, 2024.

\bibitem{MIA2020exploringDropSeg}
T.~Nair, D.~Precup, D.~L. Arnold, and T.~Arbel, ``Exploring uncertainty measures in deep networks for multiple sclerosis lesion detection and segmentation,'' \emph{Medical image analysis}, vol.~59, p. 101557, 2020.

\bibitem{UAMT2019ensembleSeg}
L.~Yu, S.~Wang, X.~Li, C.-W. Fu, and P.-A. Heng, ``Uncertainty-aware self-ensembling model for semi-supervised 3d left atrium segmentation,'' in \emph{{MICCAI}}.\hskip 1em plus 0.5em minus 0.4em\relax Springer, 2019, pp. 605--613.

\bibitem{wang2018TTA}
G.~Wang, W.~Li, S.~Ourselin, and T.~Vercauteren, ``Automatic brain tumor segmentation using convolutional neural networks with test-time augmentation,'' in \emph{International MICCAI Brainlesion Workshop}.\hskip 1em plus 0.5em minus 0.4em\relax Springer, 2018, pp. 61--72.

\bibitem{huang2022lymphoma}
L.~Huang, S.~Ruan, P.~Decazes, and T.~Denoeux, ``Lymphoma segmentation from 3d pet-ct images using a deep evidential network,'' \emph{International Journal of Approximate Reasoning}, vol. 149, pp. 39--60, 2022.

\bibitem{zou2022tbrats}
K.~Zou, X.~Yuan, X.~Shen, M.~Wang, and H.~Fu, ``Tbrats: Trusted brain tumor segmentation,'' in \emph{{MICCAI}}.\hskip 1em plus 0.5em minus 0.4em\relax Springer, 2022, pp. 503--513.

\bibitem{wang2023uncertainty}
M.~Wang, T.~Lin, L.~Wang, A.~Lin, K.~Zou, X.~Xu, Y.~Zhou, Y.~Peng, Q.~Meng, Y.~Qian \emph{et~al.}, ``Uncertainty-inspired open set learning for retinal anomaly identification,'' \emph{Nature Communications}, vol.~14, no.~1, p. 6757, 2023.

\bibitem{evidential18}
M.~Sensoy, L.~Kaplan, and M.~Kandemir, ``Evidential deep learning to quantify classification uncertainty,'' in \emph{{NeurIPS}}, 2018, pp. 3183--3193.

\bibitem{2021DetermSeg}
J.~Mukhoti, J.~van Amersfoort, P.~H. Torr, and Y.~Gal, ``Deep deterministic uncertainty for semantic segmentation,'' in \emph{International Conference on Machine Learning Workshop on Uncertainty and Robustness in Deep Learning}, 2021.

\bibitem{ICMLdeterministic20}
J.~Van~Amersfoort, L.~Smith, Y.~W. Teh, and Y.~Gal, ``Uncertainty estimation using a single deep deterministic neural network,'' in \emph{{ICML}}.\hskip 1em plus 0.5em minus 0.4em\relax PMLR, 2020, pp. 9690--9700.

\bibitem{plummer2018conditional}
B.~A. Plummer, P.~Kordas, M.~H. Kiapour, S.~Zheng, R.~Piramuthu, and S.~Lazebnik, ``Conditional image-text embedding networks,'' in \emph{{ECCV}}, 2018, pp. 249--264.

\bibitem{wang2019phrase}
J.~Wang and L.~Specia, ``Phrase localization without paired training examples,'' in \emph{{ICCV}}, 2019, pp. 4663--4672.

\bibitem{moon2022multi}
J.~H. Moon, H.~Lee, W.~Shin, Y.-H. Kim, and E.~Choi, ``Multi-modal understanding and generation for medical images and text via vision-language pre-training,'' \emph{IEEE Journal of Biomedical and Health Informatics}, vol.~26, no.~12, pp. 6070--6080, 2022.

\bibitem{wang2022multi}
F.~Wang, Y.~Zhou, S.~Wang, V.~Vardhanabhuti, and L.~Yu, ``Multi-granularity cross-modal alignment for generalized medical visual representation learning,'' \emph{{NeurIPS}}, vol.~35, pp. 33\,536--33\,549, 2022.

\bibitem{wu2023medklip}
C.~Wu, X.~Zhang, Y.~Zhang, Y.~Wang, and W.~Xie, ``Medklip: Medical knowledge enhanced language-image pre-training,'' \emph{{ICCV}}, 2023.

\bibitem{BioViL2022}
B.~Boecking, N.~Usuyama, S.~Bannur, D.~C. Castro, A.~Schwaighofer, S.~Hyland, M.~Wetscherek, T.~Naumann, A.~Nori, J.~Alvarez-Valle, H.~Poon, and O.~Oktay, ``Making the most of text semantics to improve biomedical vision--language processing,'' in \emph{ECCV}, S.~Avidan, G.~Brostow, M.~Ciss{\'e}, G.~M. Farinella, and T.~Hassner, Eds.\hskip 1em plus 0.5em minus 0.4em\relax Cham: Springer Nature Switzerland, 2022, pp. 1--21.

\bibitem{vilouras2024zero}
K.~Vilouras, P.~Sanchez, A.~Q. O'Neil, and S.~A. Tsaftaris, ``Zero-shot medical phrase grounding with off-the-shelf diffusion models,'' \emph{IEEE Journal of Biomedical and Health Informatics}, 2024.

\bibitem{wang2023chatcad}
S.~Wang, Z.~Zhao, X.~Ouyang, Q.~Wang, and D.~Shen, ``Chatcad: Interactive computer-aided diagnosis on medical image using large language models,'' \emph{arXiv preprint arXiv:2302.07257}, 2023.

\bibitem{moor2023med}
M.~Moor, Q.~Huang, S.~Wu, M.~Yasunaga, Y.~Dalmia, J.~Leskovec, C.~Zakka, E.~P. Reis, and P.~Rajpurkar, ``Med-flamingo: a multimodal medical few-shot learner,'' in \emph{Machine Learning for Health (ML4H)}.\hskip 1em plus 0.5em minus 0.4em\relax PMLR, 2023, pp. 353--367.

\bibitem{wu2023towards}
C.~Wu, X.~Zhang, Y.~Zhang, Y.~Wang, and W.~Xie, ``Towards generalist foundation model for radiology,'' \emph{arXiv preprint arXiv:2308.02463}, 2023.

\bibitem{zhao2024heterogeneous}
W.~Zhao, Z.~Deng, S.~Yadav, and P.~S. Yu, ``Heterogeneous knowledge grounding for medical question answering with retrieval augmented large language model,'' in \emph{Companion Proceedings of the ACM Web Conference 2024}, 2024, pp. 1590--1594.

\bibitem{gao2023large}
Y.~Gao, R.~Li, E.~Croxford, S.~Tesch, D.~To, J.~Caskey, B.~W. Patterson, M.~M. Churpek, T.~Miller, D.~Dligach \emph{et~al.}, ``Large language models and medical knowledge grounding for diagnosis prediction,'' \emph{medRxiv}, pp. 2023--11, 2023.

\bibitem{bannur2024maira}
S.~Bannur, K.~Bouzid, D.~C. Castro, A.~Schwaighofer, S.~Bond-Taylor, M.~Ilse, F.~P{\'e}rez-Garc{\'\i}a, V.~Salvatelli, H.~Sharma, F.~Meissen \emph{et~al.}, ``Maira-2: Grounded radiology report generation,'' \emph{arXiv preprint arXiv:2406.04449}, 2024.

\bibitem{huang2024review}
L.~Huang, S.~Ruan, Y.~Xing, and M.~Feng, ``A review of uncertainty quantification in medical image analysis: probabilistic and non-probabilistic methods,'' \emph{Medical Image Analysis}, p. 103223, 2024.

\bibitem{2018probabilisticU}
S.~Kohl, B.~Romera-Paredes, C.~Meyer, J.~De~Fauw, J.~R. Ledsam, K.~Maier-Hein, S.~Eslami, D.~Jimenez~Rezende, and O.~Ronneberger, ``A probabilistic u-net for segmentation of ambiguous images,'' \emph{{NeurIPS}}, vol.~31, 2018.

\bibitem{TMI20ensembleSeg}
A.~Mehrtash, W.~M. Wells, C.~M. Tempany, P.~Abolmaesumi, and T.~Kapur, ``Confidence calibration and predictive uncertainty estimation for deep medical image segmentation,'' \emph{IEEE Transactions on Medical Imaging}, vol.~39, no.~12, pp. 3868--3878, 2020.

\bibitem{luo2024measurement}
Y.~Luo, Q.~Yang, Y.~Fan, H.~Qi, and M.~Xia, ``Measurement guidance in diffusion models: Insight from medical image synthesis,'' \emph{IEEE Transactions on Pattern Analysis and Machine Intelligence}, 2024.

\bibitem{huang2025deep}
L.~Huang, S.~Ruan, P.~Decazes, and T.~Den{\oe}ux, ``Deep evidential fusion with uncertainty quantification and reliability learning for multimodal medical image segmentation,'' \emph{Information Fusion}, vol. 113, p. 102648, 2025.

\bibitem{lai2023lisa}
X.~Lai, Z.~Tian, Y.~Chen, Y.~Li, Y.~Yuan, S.~Liu, and J.~Jia, ``Lisa: Reasoning segmentation via large language model,'' \emph{arXiv preprint arXiv:2308.00692}, 2023.

\bibitem{hu2022lora}
E.~J. Hu, Y.~Shen, P.~Wallis, Z.~Allen-Zhu, Y.~Li, S.~Wang, L.~Wang, W.~Chen \emph{et~al.}, ``Lora: Low-rank adaptation of large language models.'' \emph{ICLR}, vol.~1, no.~2, p.~3, 2022.

\bibitem{SAM}
A.~Kirillov, E.~Mintun, N.~Ravi, H.~Mao, C.~Rolland, L.~Gustafson, T.~Xiao, S.~Whitehead, A.~C. Berg, W.-Y. Lo \emph{et~al.}, ``Segment anything,'' \emph{arXiv preprint arXiv:2304.02643}, 2023.

\bibitem{du1999centroidal}
Q.~Du, V.~Faber, and M.~Gunzburger, ``Centroidal voronoi tessellations: Applications and algorithms,'' \emph{SIAM review}, vol.~41, no.~4, pp. 637--676, 1999.

\bibitem{fastrcnn2015}
R.~Girshick, ``Fast r-cnn,'' in \emph{{ICCV}}, 2015, pp. 1440--1448.

\bibitem{giou2019}
H.~Rezatofighi, N.~Tsoi, J.~Gwak, A.~Sadeghian, I.~Reid, and S.~Savarese, ``Generalized intersection over union: A metric and a loss for bounding box regression,'' in \emph{{CVPR}}, 2019, pp. 658--666.

\bibitem{rupprecht2017Mheads}
C.~Rupprecht, I.~Laina, R.~DiPietro, M.~Baust, F.~Tombari, N.~Navab, and G.~D. Hager, ``Learning in an uncertain world: Representing ambiguity through multiple hypotheses,'' in \emph{{ICCV}}, 2017, pp. 3591--3600.

\bibitem{johnson2019mimic}
A.~E. Johnson, T.~J. Pollard, S.~J. Berkowitz, N.~R. Greenbaum, M.~P. Lungren, C.-y. Deng, R.~G. Mark, and S.~Horng, ``Mimic-cxr, a de-identified publicly available database of chest radiographs with free-text reports,'' \emph{Scientific Data}, vol.~6, no.~1, p. 317, 2019.

\bibitem{wang2017chestx}
X.~Wang, Y.~Peng, L.~Lu, Z.~Lu, M.~Bagheri, and R.~M. Summers, ``Chestx-ray8: Hospital-scale chest x-ray database and benchmarks on weakly-supervised classification and localization of common thorax diseases,'' in \emph{{CVPR}}, 2017, pp. 2097--2106.

\bibitem{hu2023expert}
X.~Hu, L.~Gu, Q.~An, M.~Zhang, L.~Liu, K.~Kobayashi, T.~Harada, R.~M. Summers, and Y.~Zhu, ``Expert knowledge-aware image difference graph representation learning for difference-aware medical visual question answering,'' in \emph{{ACM KDD}}, 2023, pp. 4156--4165.

\bibitem{devlin2019bert}
J.~Devlin, M.-W. Chang, K.~Lee, and K.~Toutanova, ``Bert: Pre-training of deep bidirectional transformers for language understanding,'' in \emph{Proceedings of the 2019 conference of the North American chapter of the association for computational linguistics: human language technologies, volume 1 (long and short papers)}, 2019, pp. 4171--4186.

\bibitem{jain2021radgraph}
S.~Jain, A.~Agrawal, A.~Saporta, S.~Q. Truong, D.~N. Duong, T.~Bui, P.~Chambon, Y.~Zhang, M.~P. Lungren, A.~Y. Ng \emph{et~al.}, ``Radgraph: Extracting clinical entities and relations from radiology reports,'' \emph{arXiv preprint arXiv:2106.14463}, 2021.

\bibitem{smit2020chexbert}
A.~Smit, S.~Jain, P.~Rajpurkar, A.~Pareek, A.~Y. Ng, and M.~P. Lungren, ``Chexbert: combining automatic labelers and expert annotations for accurate radiology report labeling using bert,'' \emph{arXiv preprint arXiv:2004.09167}, 2020.

\bibitem{RefTR2021}
L.~Muchen and S.~Leonid, ``Referring transformer: A one-step approach to multi-task visual grounding,'' in \emph{{NeurIPS}}, 2021.

\bibitem{nori2023capabilities}
H.~Nori, N.~King, S.~M. McKinney, D.~Carignan, and E.~Horvitz, ``Capabilities of gpt-4 on medical challenge problems,'' \emph{arXiv preprint arXiv:2303.13375}, 2023.

\bibitem{touvron2023llama}
H.~Touvron, T.~Lavril, G.~Izacard, X.~Martinet, M.-A. Lachaux, T.~Lacroix, B.~Rozi{\`e}re, N.~Goyal, E.~Hambro, F.~Azhar \emph{et~al.}, ``Llama: Open and efficient foundation language models,'' \emph{arXiv preprint arXiv:2302.13971}, 2023.

\bibitem{ma2024segment}
J.~Ma, Y.~He, F.~Li, L.~Han, C.~You, and B.~Wang, ``Segment anything in medical images,'' \emph{Nature Communications}, vol.~15, no.~1, p. 654, 2024.

\bibitem{liu2024gemex}
B.~Liu, K.~Zou, L.~Zhan, Z.~Lu, X.~Dong, Y.~Chen, C.~Xie, J.~Cao, X.-M. Wu, and H.~Fu, ``Gemex: A large-scale, groundable, and explainable medical vqa benchmark for chest x-ray diagnosis,'' \emph{arXiv preprint arXiv:2411.16778}, 2024.

\bibitem{wang2025enhancing}
M.~Wang, T.~Lin, A.~Lin, K.~Yu, Y.~Peng, L.~Wang, C.~Chen, K.~Zou, H.~Liang, M.~Chen \emph{et~al.}, ``Enhancing diagnostic accuracy in rare and common fundus diseases with a knowledge-rich vision-language model,'' \emph{Nature Communications}, vol.~16, no.~1, p. 5528, 2025.

\end{thebibliography}

\end{document}